\documentclass[10pt,twocolumn,letterpaper]{article}

\usepackage{iccv}
\usepackage{times}
\usepackage{epsfig}
\usepackage{graphicx}
\usepackage{amsmath}
\usepackage{amssymb}

\usepackage{booktabs}
\usepackage{subfigure}
\usepackage[table,dvipsnames]{xcolor}
\usepackage{enumitem}
\usepackage[accsupp]{axessibility}

\usepackage[pagebackref=true,breaklinks=true,letterpaper=true,colorlinks,bookmarks=false]{hyperref}

\iccvfinalcopy 

\def\iccvPaperID{6152} 

\definecolor{Gray}{gray}{0.92}
\definecolor{myblue}{rgb}{0.8313,0.9019,0.9451}
\ificcvfinal\fi

\begin{document}
	
	\title{Deep Structured Instance Graph for Distilling Object Detectors}
	
	\author{
		Yixin Chen$^{1}$, Pengguang Chen$^{1}$, Shu Liu$^{2}$, Liwei Wang$^{1}$, Jiaya Jia$^{1,2}$  \\[0.2cm]
		The Chinese University of Hong Kong$^{1}$\quad SmartMore$^{2}$
	}
	
	\maketitle
	\ificcvfinal\thispagestyle{empty}\fi

	\begin{abstract}
		Effectively structuring deep knowledge plays a pivotal role in transfer from teacher to student, especially in semantic vision tasks. In this paper, we present a simple knowledge structure to exploit and encode information inside the detection system to facilitate detector knowledge distillation. Specifically, aiming at solving the feature imbalance problem while further excavating the missing relation inside semantic instances, we design a graph whose nodes correspond to instance proposal-level features and edges represent the relation between nodes. To further refine this graph, we design an adaptive background loss weight to reduce node noise and background samples mining to prune trivial edges. We transfer the entire graph as encoded knowledge representation from teacher to student, capturing local and global information simultaneously. 
		
		We achieve new state-of-the-art results on the challenging COCO object detection task with diverse student-teacher pairs on both one- and two-stage detectors. We also experiment with instance segmentation to demonstrate robustness of our method. It is notable that distilled Faster R-CNN with ResNet18-FPN and ResNet50-FPN yields 38.68 and 41.82 Box AP respectively on the COCO benchmark, Faster R-CNN with ResNet101-FPN significantly achieves 43.38 AP, which outperforms ResNet152-FPN teacher about 0.7 AP.
		Code: \url{https://github.com/dvlab-research/Dsig}.
	\end{abstract}
	
	\section{Introduction}
	Thanks to massive visual data and computing power, there is increasing advancement of advanced object detectors driven by deep networks. The backbone networks, such as ResNet~\cite{DBLP:conf/cvpr/HeZRS16} and VGG~\cite{DBLP:journals/corr/SimonyanZ14a}, facilitate modern detectors to advance high-level vision research. These detectors are powerful and contain numerous weights. They consume considerable storage as well as computation, making it hard to be deployed on mobile devices. Parallel to previous research of network pruning~\cite{DBLP:conf/nips/HanPTD15, DBLP:journals/corr/HanMD15} and network quantization~\cite{DBLP:conf/eccv/RastegariORF16, DBLP:journals/corr/ZhouNZWWZ16, DBLP:conf/cvpr/JacobKCZTHAK18, DBLP:conf/cvpr/LiWLQYF19, DBLP:journals/corr/HanMD15}, knowledge distillation~\cite{Hinton2015DistillingTK, DBLP:conf/cvpr/ZhangXHL18, DBLP:conf/nips/LanZG18, DBLP:conf/cvpr/ParkKLC19, Zagoruyko2016PayingMA, DBLP:journals/corr/RomeroBKCGB14} transfers knowledge from the teacher model to a much smaller student model. It contributes in an effective way for network compression~\cite{DBLP:journals/corr/HanMD15}. 
	
	\begin{figure}[t]
		\begin{center}
			\subfigure[Original Image]
			{
				\includegraphics[width=0.43\linewidth,height=0.35\linewidth]
				{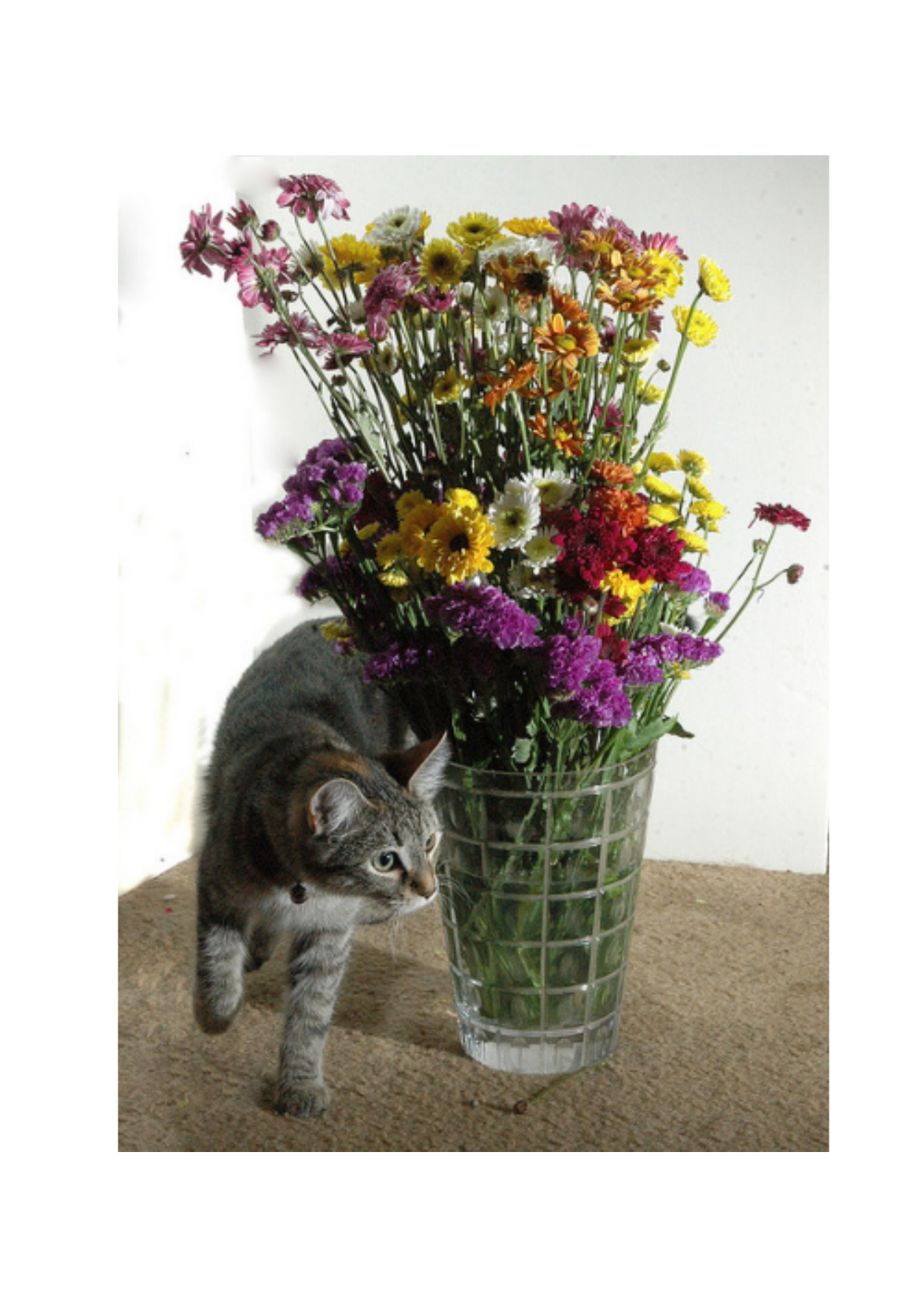}
			}\hspace{0.4cm}
			\subfigure[Ground Truth]{
				\includegraphics[width=0.43\linewidth,height=0.35\linewidth]{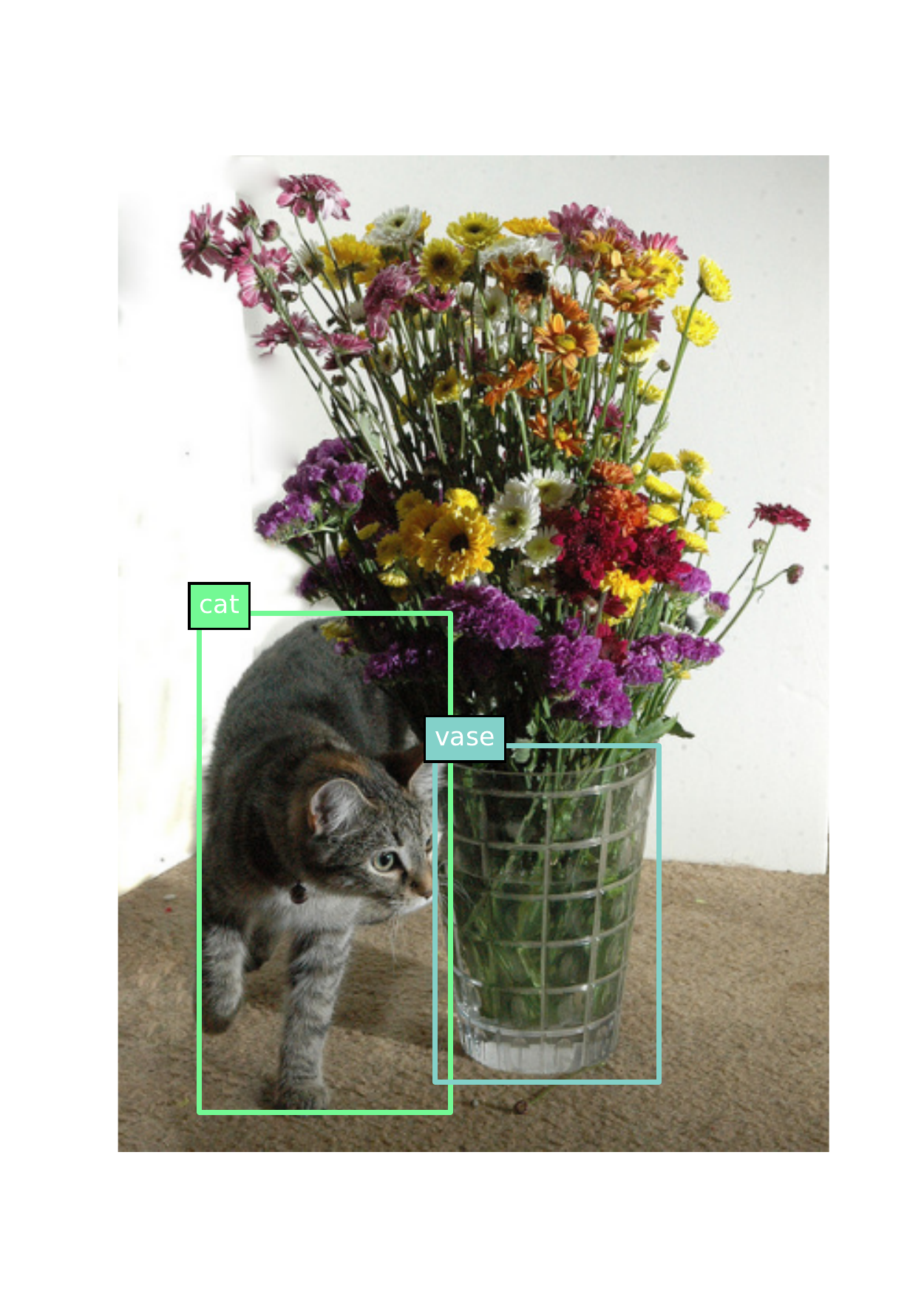}
			}\vspace{-0.25cm}
			\subfigure[student]{
				\centering
				\includegraphics[width=0.47\linewidth]{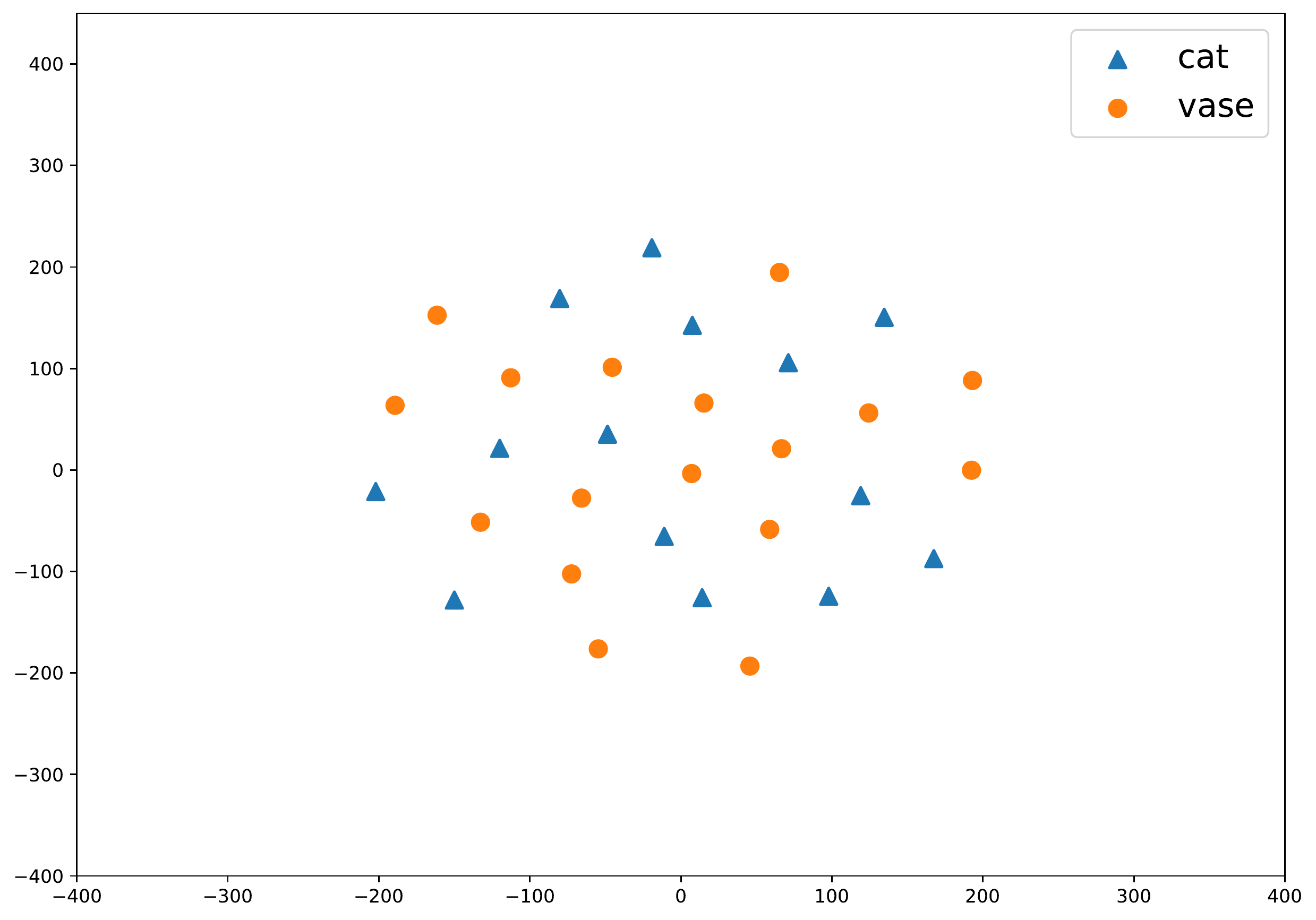}
			}
			\subfigure[teacher]{
				\centering
				\includegraphics[width=0.47\linewidth]{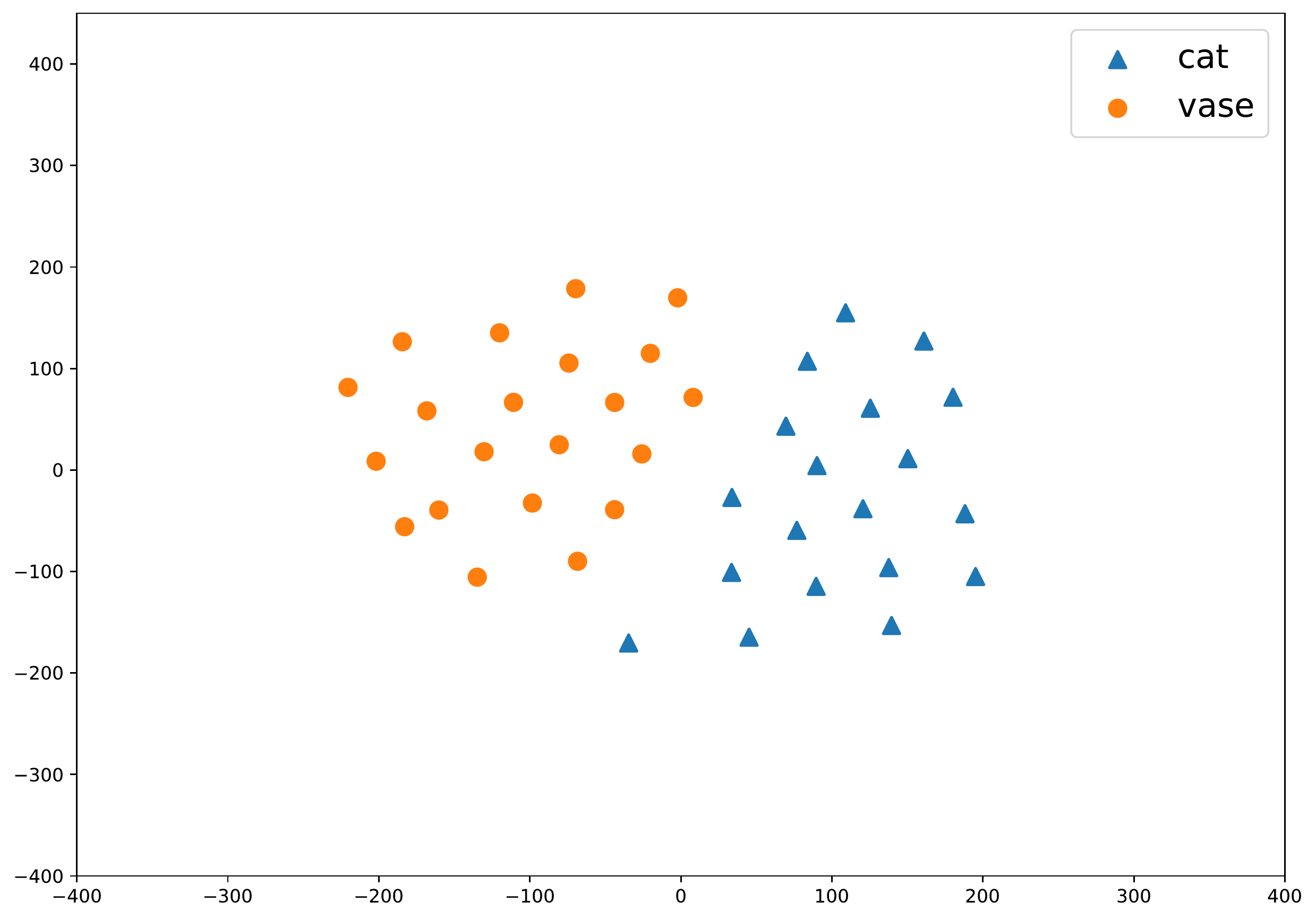}
			}\vspace{-0.25cm}
			\subfigure[pixel matching method~\cite{DBLP:conf/cvpr/WangYZF19}]{
				\centering
				\includegraphics[width=0.47\linewidth]{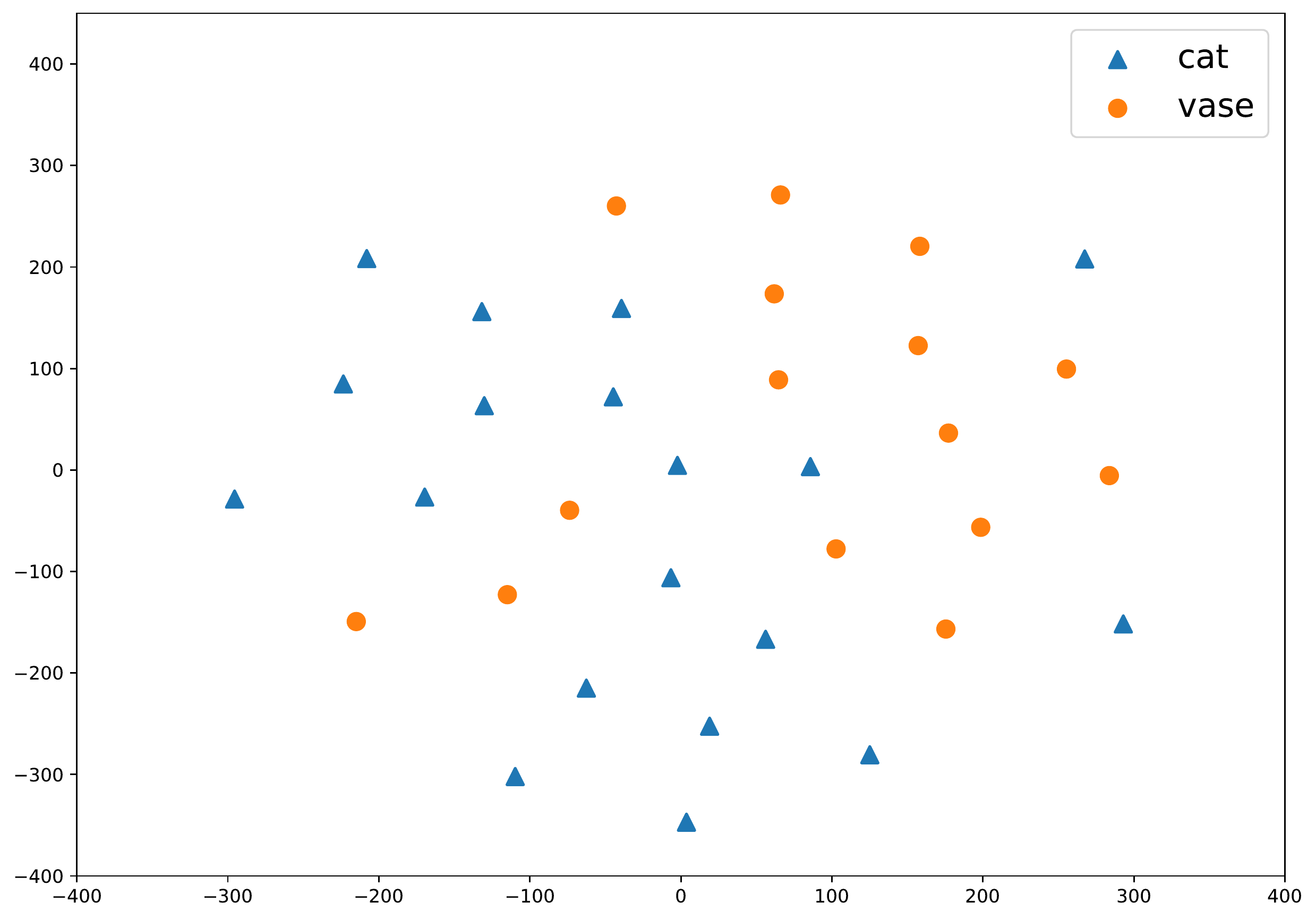}
			}
			\subfigure[ours with relation distillation]{\label{fig:ourswithrelkd}
				\centering
				\includegraphics[width=0.47\linewidth]{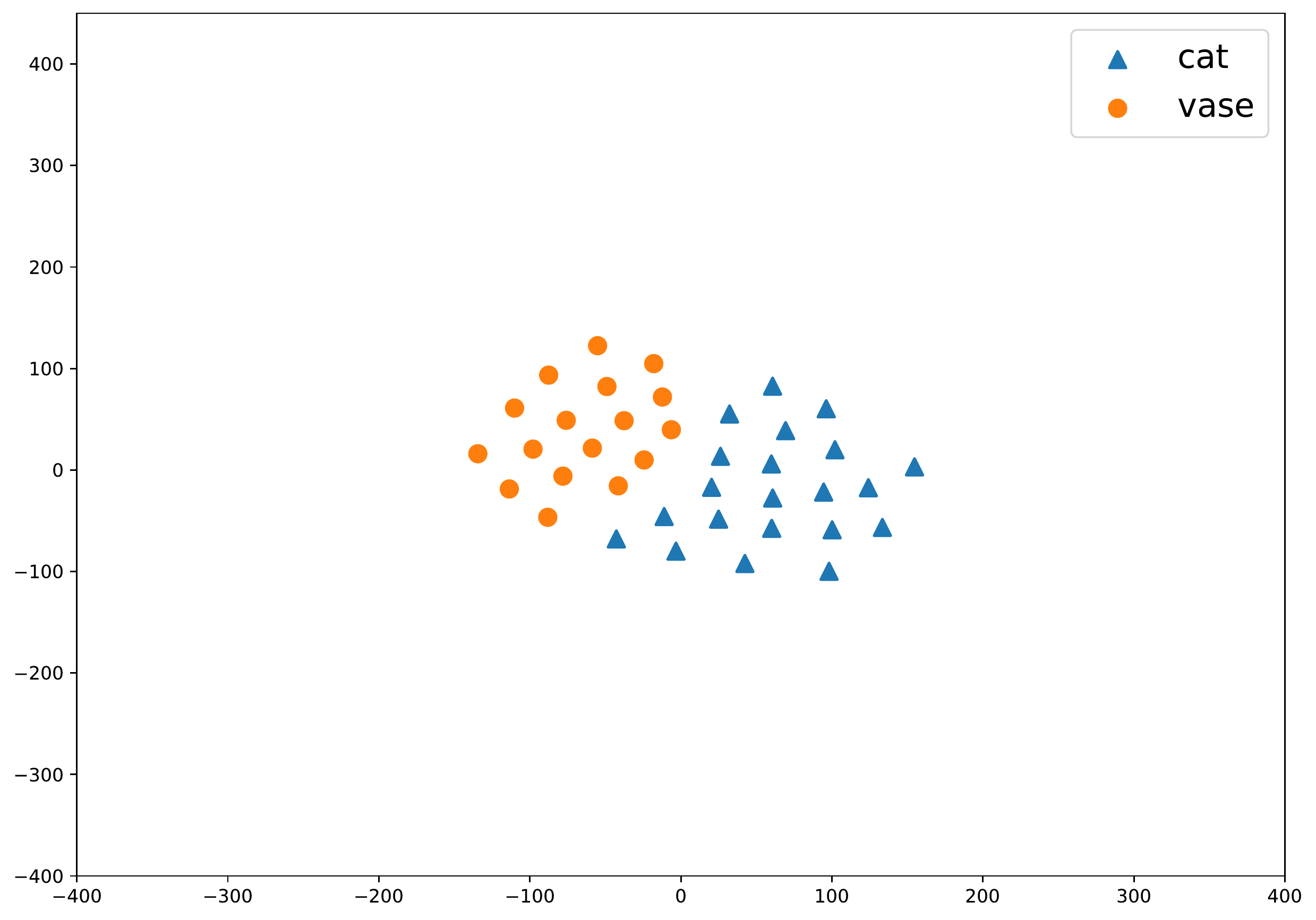}
			}
		\end{center}
		\vspace{-0.1in}
		\caption{We use t-SNE~\cite{vanDerMaaten2008} to show the topological structure of the proposal's features in different trained detectors on test image. Each marker represents one proposal's features.}
		\vspace{-0.1in}
		\label{fig:structviz}
	\end{figure}
	
	\vspace{-0.1in}
	\paragraph{Feature Imbalance:} Methods of~\cite{Hinton2015DistillingTK, DBLP:journals/corr/RomeroBKCGB14} for knowledge distillation mostly dedicate to classifier distillation where only the logits~(to the final softmax layer) are considered. 
	However, transferring large global feature maps from the teacher to student needs global feature regression, and may introduces many trivial pixels to match. 
	
	To distill useful information in feature maps, methods of~\cite{DBLP:conf/cvpr/WangYZF19, DBLP:journals/corr/abs-2006-13108} pay attention to foreground location and use human-made masks to extract close-to-instance features, leaving a level of pixels unused in the whole feature maps. Consequently, covering masks on feature maps may cause very few background features distilled, still losing useful information in distillation. These two extreme cases raise an essential question: \textit{how to leverage background features and reach a promising balance?}
	
	\vspace{-0.1in}
	\paragraph{Missing Instance-level Relations:} Additionally, all previous methods~\cite{DBLP:conf/cvpr/LiJY17,DBLP:conf/cvpr/WangYZF19, DBLP:journals/corr/abs-2006-13108} adopt the scheme to individually transfer knowledge from teacher features to the student in pixel level. In fact, object instances in a single image show latent relation~\cite{DBLP:conf/cvpr/HuGZDW18, DBLP:conf/cvpr/Liu0SC18} among each other, which is important for the sampled instance features to form knowledge base to facilitate later classification and regression tasks. 
	
	To better understand the relation, we visualize it using t-SNE~\cite{vanDerMaaten2008}, which depicts the different topological structure of instances in trained models in Figure~\ref{fig:structviz}. It reveals that the relation space of the student and teacher is quite different in terms of both shape and intensity \wrt the same test image. Moreover, after the student is distilled with only pixel-to-pixel regression~\cite{DBLP:conf/cvpr/WangYZF19}, the topological structure is no longer aligned with the teacher though it looks like better classified than the student baseline. Here thus comes another major question: \textit{how to better utilize the latent relation inside deep neural networks?} 
	
	\vspace{-0.1in}
	\paragraph{Our Contributions:} We address these two problems and define an effective \textbf{structured instance graph} based on each Region of Interest~(RoI) in the detection system. In our graph, nodes correspond to the features of RoI instances, we collect these regional features that are sampled in the subsequent classification and regression tasks. 
	
	Edges represent the relations between nodes and are measured by their feature similarity. As the architectures of student/teacher are heterogeneous in width and depth, their output is with different topological structure, shedding light on pairwise interrelation distillation. Different from pixel-to-pixel distillation, pairwise interrelation distillation utilizes information within a number of instances and introduces a new type of regularization for student training. 
	
	The nodes are devised to overcome the feature imbalance problem and the edges excavate the missing instance relation. Rather than transferring the nodes and edges separately, we directly distill the structured graph from teacher to student via a simple loss function, to close the gap between their knowledge space. In Figure~\ref{fig:ourswithrelkd}, intuitively, distilling the entire graph via our method is actually to match local feature patches while capturing the global topological structures in the meantime. 
	
	However, distilling the graph is not easy. First, a large proportion of background nodes in distillation provide too much noisy supervision compared with foreground nodes. Second, dense connection between nodes also contains massive background-related edges~(linked with background node). These two issues both add harmful regularization to overwhelm the distillation process. Here we introduce two techniques. For nodes, we control the background node loss as adaptive concerning the foreground/background ratios. For edges, we design the Background Samples Mining approach to prune trivial background-related edges, which propels remaining ambiguous false negatives to be well regularized in distillation. More details are in Section \ref{section:method}.  
	
	Our method is easy to implement and can be stably trained in the one/two-stage detection system without any additional training strategies and tricks. In experiments, our method outperforms all previous state-of-the-art detector distillation methods and achieves decent performance on the COCO detection task~\cite{DBLP:conf/eccv/LinMBHPRDZ14} regarding various student-teacher pairs. Also, we have validated our method on the COCO instance segmentation task to emphasize that our method is a general distillation framework.
	
	\section{Related Work}
	\begin{figure*}[t]
		\begin{center}
			\includegraphics[width=1.0\linewidth, height=0.35\linewidth]{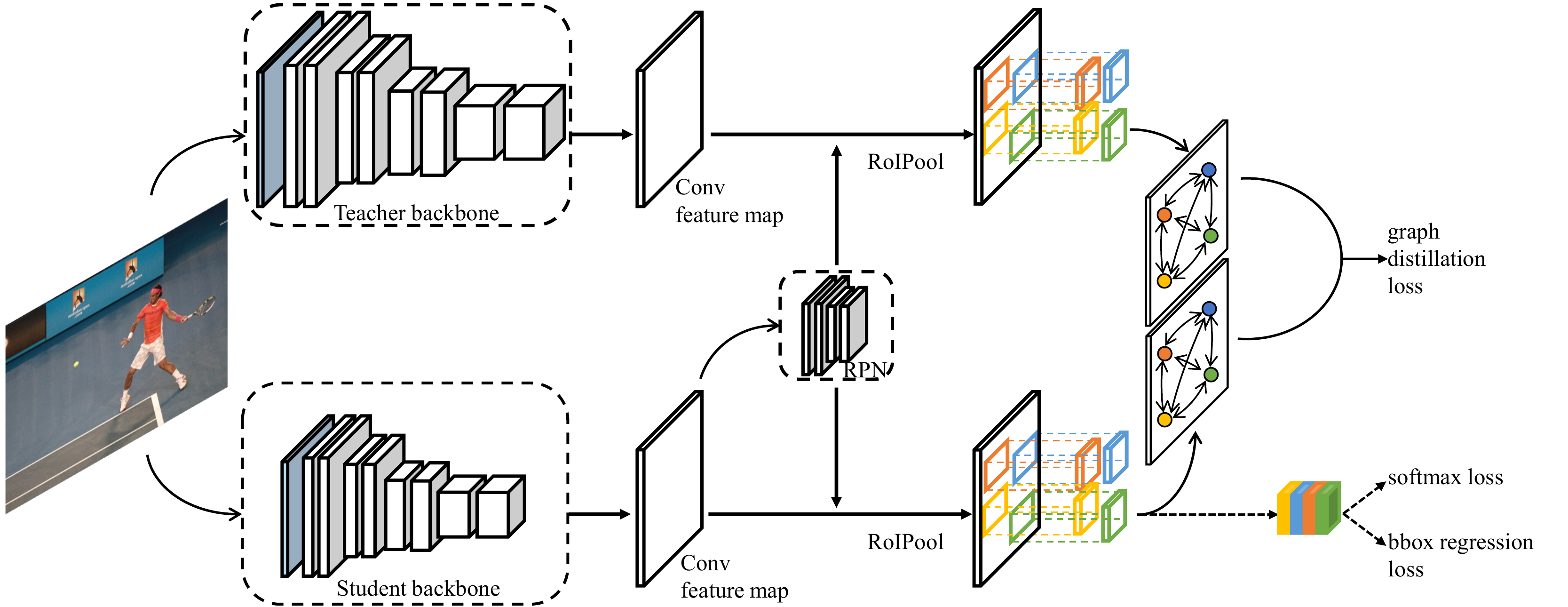}
		\end{center}
		\caption{Diagram of our method for the distillation framework. Note we share the student's RoI with teacher.}
		\label{fig:diagram}
		\vspace{-0.2in}
	\end{figure*}
	
	\subsection{Object Detectors}
	Modern CNN-based object detectors are grouped into two families according to their detection pipelines: (1) two-stage object detectors with regional proposals method; (2) one-stage object detectors with no prior proposals.
	
	Two-stage object detectors mainly derive from R-CNN~\cite{DBLP:conf/cvpr/GirshickDDM14} approach, which manages a number of candidate object regions and forward each of them independently to classify object instances and refine bounding boxes. To reduce the computational cost, SPP~\cite{DBLP:conf/eccv/HeZR014} and Fast R-CNN~\cite{girshickICCV15fastrcnn} identify RoIs on feature maps adopting RoIPool to achieve fast speed and high accuracy. Faster R-CNN~\cite{renNIPS15fasterrcnn} refined this procedure by replacing proposals generation with learnable proposals generation module Region Proposal Network~(RPN). It was the leading framework for advanced detectors~\cite{DBLP:conf/iccv/HeGDG17, DBLP:conf/nips/DaiLHS16, DBLP:conf/cvpr/0008CYLWL020, Cai_2019}. 
	
	More recently, one-stage object detectors~\cite{lin2017focal, DBLP:conf/iccv/TianSCH19, DBLP:conf/cvpr/RedmonDGF16, Liu_2016} were proposed for real-time detection while achieving considerable accuracy. In this paper, we consider distilling both one- and two-stage object detectors to show the generality of our work.

	\subsection{Deep Knowledge Representation}
	Encoding and managing knowledge in deep neural networks are of vital importance in knowledge passing between teacher and student. Hinton~\etal~\cite{Hinton2015DistillingTK} regarded the soft prediction logits as dark knowledge and matched them in distillation. Besides the logits produced by the last layer, Romero~\etal~\cite{DBLP:journals/corr/RomeroBKCGB14} proposed that intermediate representations learned by teachers as hints can also serve as a form of knowledge to improve student's performance. Zagoruyko~\etal~\cite{Zagoruyko2016PayingMA} leveraged the attention maps to guide student. Recently, instead of using individual data examples, Park~\etal~\cite{DBLP:conf/cvpr/ParkKLC19} introduced relation of image instances as a kind of knowledge transferred from teacher to student in classification. Liu~\etal~\cite{DBLP:conf/cvpr/LiuCLQLW19} utilized pixel relations in large network feature maps to facilitate student for semantic segmentation. 
	
	However, there exists no previous work to manage knowledge in a structural form in distillation for a 2D object detector. We also found in a single image, the regional instances reveal more structured semantic correlations between each than classification or semantic segmentation. In this paper, we build our graph edges based on the relation of RoI instances as deep knowledge for distillation.

	\subsection{Detector Distillation}
	Distilling knowledge from large teacher detectors to student is now an active research topic. Chen~\etal~\cite{DBLP:conf/nips/ChenCYHC17} proposed an end-to-end trainable framework for distilling multi-class object detectors. 
	Li~\etal~\cite{DBLP:conf/cvpr/LiJY17} matched all features based on region proposals.
	Recently, Wang~\etal~\cite{DBLP:conf/cvpr/WangYZF19} utilized fine-grained imitation masks to distill the near-object regions of feature maps for distillation. Sun~\etal~\cite{DBLP:journals/corr/abs-2006-13108} presented a task adaptive distillation framework with the decay strategy to improve model generalization. All of them do not elaborately employ background features. 
	
	Zhang~\etal~\cite{zhang2021improve} proposed an attention-guided method to distill useful information and introduced non-local module~\cite{DBLP:conf/cvpr/0004GGH18} to capture relation inside pixels of backbone feature maps. They ignore the inner structure inside semantic instances. In contrast, our method designs a structured graph that leverages both feature and inter-feature similarity. It transfers knowledge in a structured manner, which makes it possible to improve detector distillation effectively.

	\section{Our Method} \label{section:method}
	In this section, we introduce the distillation framework. The core idea is to generate a deep structured instance graph inside both teacher and student, based on regional object instances. This graph well exploits the deep knowledge inside detection networks and can be regarded as a new knowledge structure encoded in the detection system. Distilling the structured graph enables not only sufficient knowledge passing but also retains the whole topological structure of the embedding space.

	\subsection{Structured Instance Graph}
	Our diagram is shown in Figure~\ref{fig:diagram}. It can be applied to one- and two-stage detection networks. For illustration, we choose the classical detection network Faster R-CNN~\cite{renNIPS15fasterrcnn} for explanation. As for the one-stage detector, we can simply replace the RPN with the predicted boxes and build our graph. Unlike other methods processing the whole backbone feature map, we pay our attention to building graphs upon RoI pooled features since they are extracted based on the RPN proposals and forwarded to the subsequent detection head. Moreover, they are semantic instances that are identified by detectors. Thus, it is reasonable to model relations between instances other than independent pixels~\cite{DBLP:conf/cvpr/LiuCLQLW19}.
	
	In the structured graph, each \textbf{node} corresponds to one instance in an image, represented as the vectorized feature of this instance.
	The relation of two instances forms the \textbf{edge} between two corresponding nodes and is calculated by their similarity in the embedding space. In fact, the definition and semantics of \textbf{edge} are fundamentally different from pixel similarity in~\cite{DBLP:conf/cvpr/LiuCLQLW19}. It is notable that our nodes are pooled and extracted by the learnable semantic proposals of various scales and sizes, thus sharing strong semantic relation between each other. While those in~\cite{DBLP:conf/cvpr/LiuCLQLW19} are sampled and uniformly-distributed pixel blocks with the same sizes within an image. The strong relation between instances transferred from teacher to student would serve for interpretable distillation in our method. 
	
	Note that we share the student's RoIs with teacher to align their sampled regions. It means the same RoIs are used to extract features of student and teacher. For teacher $t$ and student $s$, the structured graph is expressed as $\mathcal{G}^t = (\mathcal{V}^t, \mathcal{E}^t)$. $\mathcal{G}^s = (\mathcal{V}^s, \mathcal{E}^s)$ can be obtained similarly, where $\mathcal{V}$ and $\mathcal{E}$ denote the node and edge sets of each graph. More definitions of nodes and edges are as follows.

	\subsubsection{Nodes}
	We directly construct nodes based on RoI pooled features. They are assigned to foreground categories or background as per IoUs between proposals and ground truth boxes. Different from previous work, we recognize that the background-labeled region features can influence the detector performance significantly. Rather than discarding these background-labeled nodes whose IoU with ground-truth boxes are less than a threshold (\eg 0.5) to avoid background noise, we divide these nodes into foreground and background in the nodes set and deal with the types differently via adaptive background loss weight~(Section \ref{sec:graphkdloss}).     
	
	The nodes in $\mathcal{G}$ are denoted as $\mathcal{V}=\{v^{fg}_{1}, v^{fg}_{2}, ..., v^{fg}_{n}, v^{bg}_{1}, v^{bg}_{2}, ..., v^{bg}_{m}\}$, where $v^{fg}_{i}$ is the feature of $i$-th foreground instance $x^{fg}_{i}$ while $v^{bg}_{i}$ is the feature of $i$-th background instance $x^{bg}_{i}$. The numbers of foreground and background instances are $n$ and $m$ respectively. Note that $n$ and $m$ vary in each image.

	\subsubsection{Edges}
	The edges in $\mathcal{G}$ are denoted as 
	$\mathcal{E} = [e_{ij}]_{k \times k}$,
	where $k$ is the size of nodes set. $e_{pq}$ is the edge of the $p$-th and $q$-th nodes, denoting the similarity of corresponding instances in the embedding space and expressed as 
	\begin{equation}\label{eq:edge}
		e_{pq} :=  \textrm{sim\_function}(v^{}_{p}, v^{}_{q}),
	\end{equation}
	where $v^{}_{i}$ denotes the node of the $i$-th instance $x^{}_{i}$.
	Here we adopt cosine similarity to define the edges of
	$$s(v^{}_p, v^{}_q)=\frac{v^{}_p\cdot v^{}_q}{\|v^{}_p\|\cdot \|v^{}_q\|},$$ 
	because it is invariable to the length of feature in $\mathcal{V}$.
	Obviously, $\mathcal{G}$ is a complete graph, since we assume that between every pair of nodes in $\mathcal{V}$ there exists an edge. Further, since the similarity function is symmetric, ${e_{pq}} = {e_{qp}}$ for any $p$ and $q$, making $\mathcal{G}$ an undirected graph and $\mathcal{E}$ a symmetric matrix with elements all being \textbf{1} in principal diagonal.

	\subsubsection{Background Samples Mining}
	We discover that distilling dense edges produced by the whole nodes set can be detrimental to training because a large amount of background nodes bring overwhelming loss in background-related edge distillation. A simple way of moderating this degeneration is to establish a smaller edge set with only foreground nodes.
	
	However, pruning all background-related edges loses too much information at the beginning of training since some of them are hard negative samples that are quite informative in training. So we design a method, called {\it Background Samples Mining} to select eligible background nodes along with the entire foreground nodes to construct edges. Assuming the original edge set based on only $n$ foreground nodes is 
	$\mathcal{E} = [e_{ij}]_{n \times n}$,
	we expand it to 
	$\mathcal{\hat{E}} = [e_{ij}]_{\hat{n} \times \hat{n}}$
	with more node links from $n \times n$ to $\hat{n} \times \hat{n}$, which means we mine $\hat{n}-n$ samples from background-labeled ones. 
	
	Inspired by OHEM~\cite{DBLP:conf/cvpr/ShrivastavaGG16}, here we introduce a technique to mine part of qualified background samples whose classification losses in teacher are greater than a threshold $T$. It intuitively reveals that these background samples are prone to misclassification, and thus can be reasonably added to the foreground-only edges set~(note all edges are linked with foreground nodes), which still establishes a dense graph. 
	
	Samples with high confidence to be classified to background are not added to the set. It is natural that the expanded edges set $\mathcal{\hat{E}}$ degenerates to the prototype $\mathcal{E}$ if no samples are mined. We also provide detailed pseudo algorithms of background samples mining and graph establishment in supplementary material.

	\subsection{Graph Distillation Loss} \label{sec:graphkdloss}
	The graph distillation loss $L_\mathcal{G}$ is defined as the discrepancy between structured graphs of teacher and student, consisting of graph node loss $L_\mathcal{V}$ and graph edge loss $L_\mathcal{E}$. We simply utilize the Euclidean distance function to evaluate these two losses as
	\begin{equation}\label{eq:lossg}
		\begin{split}
			L_\mathcal{G}\! &=\! \lambda_1 \cdot L_{\mathcal{V}}^{fg}\! +\! \lambda_2 \cdot L_{\mathcal{V}}^{bg}\! +\! \lambda_3 \cdot L_\mathcal{E}     \\
			\!&=\! \frac{\lambda_1}{N_{fg}} \!\sum_{i=1}^{N_{fg}}{\|v^{t,fg}_{i}-v^{s,fg}_{i}\|^2}\! +\! \frac{\lambda_2}{N_{bg}} \!\sum_{i=1}^{N_{bg}}{\|v^{t,bg}_{i}-v^{s,bg}_{i}\|^2} \\
			& \ \quad\! +\! \frac{\lambda_3}{N^2} \sum_{i=1}^N{\sum_{j=1}^N{\|e^{t}_{ij}-e^{s}_{ij}\|^2}}
		\end{split}
	\end{equation}
	where $\lambda_1$, $\lambda_2$, and $\lambda_3$ represent the penalty coefficient balanced in graph distillation loss. 
	We set $\lambda_1$ and $\lambda_3$ to 0.5 based on grid search on the validation set, and define $\lambda_2$ as an adaptive loss weight for background nodes to mitigate the imbalanced problem, expressed as
	\begin{equation} \label{eq:lambda2}
		\lambda_{2} = \alpha \cdot \frac{N_{fg}}{N_{bg}},
	\end{equation}
	where the $\alpha$ is a coefficient empirically set to achieve a loss scale comparable with other distillation losses.  
	
	The graph node loss $L_\mathcal{V}$ is the imitation loss between node set, it basically aligns student instance features with those of teacher in a pixel-to-pixel manner. Traditionally, directly matching the feature map between two networks is popular in distillation. However, in detection models, not all the pixels in feature maps are forwarded to produce the classification and box regression loss. Rather than utilizing the overall feature map, we adopt the sampled foreground and background features to produce the graph node loss. It pushes the student to focus more on the RoIs along with useful knowledge. 
	
	The graph edge loss $L_\mathcal{E}$ is the imitation loss between edges set. It leads to relation of student node alignment with those of teacher. In experiments, simply mimicking features cannot thoroughly mine the potential of knowledge. When the highly semantic relation is not well distilled with the node loss, edge loss would otherwise directly propel the pairwise interrelation that is to be learned. Therefore, to match the topological knowledge space between the student and teacher, it is necessary to design the edge loss to capture the global structured information in detectors.

	\subsection{Overall Loss}
	It is common in image classification to transfer knowledge from teacher logits to student ones~\cite{Hinton2015DistillingTK}. In detection, we have our classification and bounding box head, in which the output logits are matched using Kullback-Leibler~(KL) Divergence loss. A detailed definition of KLD loss is given in supplementary material.
	
	Incorporating graph and head logits KLD loss into the detector loss, we form the overall student training loss as
	\begin{equation}\label{eq:lossoverall}
		\begin{split}
			L &= L_{Det} + L_\mathcal{G} + L_{Logits} \\
			&= L_{RPN} + L_{RoIcls} + L_{RoIreg} \\ 
			& \ \quad + L_\mathcal{G} + L_{Logits}
		\end{split}
	\end{equation}
	where $L_{RPN}$, $L_{RoIcls}$, and $L_{RoIreg}$ represent the supervised RPN loss, RoI classification loss, and RoI bounding box regression loss, $L_{Logits}$ represents the classification and bbox regression logits KLD loss. Moreover, $\lambda_1$,  $\lambda_3$, and $\alpha$ in $L_\mathcal{G}$~(Eq~\eqref{eq:lossg}) are kept unchanged during training. 

	\section{Experiments}
	
	\paragraph{Experimental Benchmark} We adopt the challenging object detection benchmark COCO~\cite{DBLP:conf/eccv/LinMBHPRDZ14} to validate the effectiveness of our proposed method. Following the common practice, we train and validate all our COCO models on train/val2017, which contains around 118k/5k images respectively. For evaluation, the detection average precision~(AP) over IoU threshold is adopted, and we report our results on COCO style AP metrics including AP@[0.5:0.95], AP$_{50}$, AP$_{75}$, AP$_{S}$, AP$_{M}$, and AP$_{L}$.    
	
	\vspace{-0.1in}
	\paragraph{Network Architecture and Initialization} 
	We build our experiment upon Detectron2~\cite{wu2019detectron2}, and adopt the off-the-shelf pre-trained Detectron2 model zoo
	as teachers. In details, different sizes and architectures of backbone act as teacher and student. We choose ResNet-FPN\cite{DBLP:conf/cvpr/LinDGHHB17}-3x\footnote{ResNet-FPN-3x: ResNet-FPN as backbone and train for 3x schedule.} as teacher architecture. 
	
	Apart from ResNets, we also adopt MobileNetV2~\cite{DBLP:conf/cvpr/SandlerHZZC18} and EfficientNet-B0~\cite{DBLP:conf/icml/TanL19} as backbones for student. We further evaluate our method on one-stage detectors of RetinaNet~\cite{lin2017focal} with these backbones. Note that 1x/3x schedule in COCO means around 12/37-epoch training.     
	
	\vspace{-0.1in}
	\paragraph{Training Details}\label{section:training_detail} 
	With the supervision of pre-trained teacher models, we train students detection networks with different types of architecture and capacity. We conduct experiments on multiple student-teacher pairs of R18-R50, R50-R101, R101-R152, MNV2-R50, and EB0-R101, to verify our method. For training, all our experiments are performed on 4 Nvidia RTX 2080Ti GPUs, and all students stick to the 1x/2x/3x COCO training schedule.
	Detailed training setting is provided in the supplementary material.
	
	\subsection{Main Results} 

	We present our overall distillation performance of two-stage detector Faster R-CNN as well as one-stage detector RetinaNet for multiple student-teacher~(Section \ref{section:training_detail}) on the COCO dataset~\cite{DBLP:conf/eccv/LinMBHPRDZ14}. For Faster R-CNN, as shown in Table~\ref{table:cocoperformance}, 
	student R18 improves its baseline by 4.19 AP, with larger capacity, student R50/R101 still surpasses the baseline of 2.54/1.38 AP, which proves the robustness of our method even when the gap between student-teacher varies.
	
	We also evaluate distilling detectors with heterogeneous student-teacher (EB0-R101, MNV2-R50). Despite distilled by totally different architectures, student EB0 and MNV2 still gets considerable AP gain (3.89/4.97), which manifests that our graph can be effectively adopted in diversified types of backbones. For RetinaNet, it is observed in Table~\ref{table:onestageperformance} that all students achieve stable gain \wrt baseline, which shows that our method is generative for one-stage detector too.

	Since 1x models are heavily under-trained, we also provide sufficiently trained 3x models results, see Table~\ref{table:cocoperformance}~\ref{table:onestageperformance}.
	For two-stage Faster-RCNN, 3x-distilled models achieve substantial 5.3 AP promotion on average, and some of them even outperform the teacher by large margins. For one-stage detector, there is 5.49 average AP improvement on 3x-distilled RetinaNet. These student-teacher-3x pairs all yield satisfactory results, indicating that our method is applicable when training is even longer.   
	
	\begin{table}[t]
		\begin{center}
			\renewcommand\arraystretch{0.98}
			\begin{tabular}{c|cc|c|c}
				\toprule
				Detector & Student & Teacher & Schedule & AP$_{box}$   \\
				\midrule \toprule
				Faster RCNN & R18 & -    & 1x & 33.06  \\
				Faster RCNN &R18 & R50  & 1x & \textbf{37.25} \\
				Faster RCNN &R18 & R50  & 3x & \textbf{38.68} \\
				Faster RCNN &- & R50    & 3x & 40.22 \\
				\midrule
				Faster RCNN &R50 & -    & 1x & 38.03 \\
				Faster RCNN &R50 & R101 & 1x & \textbf{40.57} \\
				Faster RCNN &R50 & R101 & 3x & \textbf{41.82} \\
				Faster RCNN &- & R101   & 3x & 42.03 \\
				\midrule
				Faster RCNN &R101 & -    & 1x & 40.27 \\
				Faster RCNN &R101 & R152   & 1x & \textbf{41.65} \\
				Faster RCNN &R101 & R152   & 3x & \textbf{43.38} \\
				Faster RCNN &- & R152   & 3x & 42.66 \\
				\midrule
				Faster RCNN &EB0 & -    & 1x &  33.85 \\
				Faster RCNN &EB0 & R101  & 1x & \textbf{37.74} \\
				Faster RCNN &EB0 & R101  & 3x & \textbf{40.39} \\
				Faster RCNN &- & R101    & 3x & 42.03 \\
				\midrule
				Faster RCNN &MNV2 & -    & 1x & 29.47   \\
				Faster RCNN &	MNV2 & R50  & 1x & \textbf{34.44}  \\
				Faster RCNN &MNV2 & R50  & 3x &  \textbf{36.93} \\
				Faster RCNN &- & R50    & 3x & 40.22 \\
				\bottomrule
			\end{tabular}
		\end{center}
		\caption{Object detection Box AP on COCO2017 \texttt{val} using Faster R-CNN with various backbones of ResNet18(\textbf{R18}), ResNet50(\textbf{R50}), ResNet101(\textbf{R101}), EfficientNetB0(\textbf{EB0}), and MobileNetV2(\textbf{MNV2}). Note that the \textit{dash} refers to ``none student or teacher exists", a student and teacher baseline.}
		\label{table:cocoperformance}
	\end{table}

	\begin{table}[t]
		\begin{center}
			\renewcommand\arraystretch{0.98}
			\begin{tabular}{c|cc|c|c}
				\toprule
				Detector & Student & Teacher & Schedule & AP$_{box}$   \\
				\midrule \toprule
				RetinaNet & R18 & -    & 1x & 31.60 \\
				RetinaNet & R18 & R50  & 1x & \textbf{34.72} \\
				RetinaNet & R18 & R50  & 3x &  \textbf{37.18}\\
				RetinaNet & - & R50    & 3x & 38.67 \\
				\midrule
				RetinaNet & MNV2 & -    & 1x &  29.31 \\
				RetinaNet & MNV2 & R50  & 1x & \textbf{32.16} \\
				RetinaNet & MNV2 & R50  & 3x & \textbf{35.70}\\
				RetinaNet & -    & R50  & 3x &  38.67 \\
				\midrule
				RetinaNet & EB0 & -    & 1x & 33.35\\
				RetinaNet & EB0 & R101  & 1x &  \textbf{34.44}\\
				RetinaNet & EB0 & R101  & 3x &  \textbf{37.86}\\
				RetinaNet & -    & R101  & 3x & 40.39 \\
				\bottomrule
			\end{tabular}
		\end{center}
		\caption{Object detection Box AP on COCO2017 \texttt{val} using One-Stage Detector RetinaNet with various backbones.}
		\vspace{-0.2in}
		\label{table:onestageperformance}
	\end{table}


	\subsection{Comparison with other methods}
	
	\begin{table}[t]
		\begin{center}
			\renewcommand\arraystretch{0.98}
			\setlength\tabcolsep{2pt}
			\begin{tabular}{c|c|c|cc}
				\toprule
				Method & Stu-Tch & Schedule & AP$_{box}$ & AP$_{mask}$ \\
				\midrule \toprule
				Stu Baseline&R18    & 1x & 33.89 & 31.30  \\
				$\dagger$PixelPairWise~\cite{DBLP:conf/cvpr/LiuCLQLW19} 
				& R18-50 & 1x & 33.63 & 30.43 \\
				$\dagger$FGFI~\cite{DBLP:conf/cvpr/WangYZF19} 
				& R18-50 & 1x & 34.39& 31.49 \\
				\multicolumn{1}{>{\columncolor{Gray}[2pt][149pt]}c|}{Ours}
				& R18-50  & 1x & \textbf{37.33} & \textbf{33.90}  \\
				\multicolumn{1}{>{\columncolor{Gray}[2pt][149pt]}c|}{Ours}
				& R18-50  & 3x & \textbf{39.05} & \textbf{35.49} \\
				Tch Baseline& R50  & 3x & 40.98 & 37.16  \\
				\midrule
				Stu Baseline & R50  & 1x & 38.64 & 35.24  \\
				$\dagger$PixelPairWise~\cite{DBLP:conf/cvpr/LiuCLQLW19} 
				& R50-101 & 1x & 38.80& 34.89 \\
				$\dagger$FGFI~\cite{DBLP:conf/cvpr/WangYZF19} 
				& R50-101 & 1x &38.97 & 35.30 \\
				\multicolumn{1}{>{\columncolor{Gray}[2pt][149pt]}c|}{Ours}
				&R50-101 & 1x & \textbf{40.06} & \textbf{36.28}  \\
				AttentionGuided~\cite{zhang2021improve} 
				& R50-101 & 2x & 41.70 & 37.40 \\
				\multicolumn{1}{>{\columncolor{Gray}[2pt][149pt]}c|}{Ours}
				&R50-101 & 2x & 41.64& 37.52   \\
				\multicolumn{1}{>{\columncolor{Gray}[2pt][149pt]}c|}{Ours}
				&R50-101 & 3x & \textbf{42.23} & \textbf{38.06} \\
				Tch Baseline& R101 & 3x & 42.92 & 38.63  \\
				\bottomrule
			\end{tabular}
		\end{center}
		\caption{Instance segmentation results AP on COCO2017 \texttt{val} using Mask R-CNN with ResNet backbones. \textbf{Stu} and \textbf{Tch} refers to student and teacher respectively. $\dagger$Methods are reproduced by ourselves, other results are obtained from corresponding papers.} 
		\vspace{-0.2in}
		\label{table:cocoperformance_instanceseg}
	\end{table}
	
	\begin{table*}[t]
		\begin{center}
			\renewcommand\arraystretch{0.98}
			\begin{tabular}{c|c|c|c|cccccc}
				\toprule
				Detector &Method & BackBone & Schedule & AP & AP$_{50}$ & AP$_{75}$ & AP$_S$ & AP$_M$ & AP$_L$  \\
				\midrule \toprule
				Faster RCNN &Student Baseline & ResNet18 & 1x  & 33.06 & 53.43 & 35.19 & 18.83 & 35.64 & 42.73 \\
				Faster RCNN &Teacher Baseline & ResNet50 & 3x  & 40.22 & 61.01 & 43.81 & 24.15 & 43.52 & 51.97 \\
				\midrule
				Faster RCNN & $\dagger$ FGFI~\cite{DBLP:conf/cvpr/WangYZF19} & ResNet18 &  1x  & 34.16 & 54.25 & 36.70 & 18.79 & 36.92 & 44.73 \\
				Faster RCNN &$\dagger$PixelPairWise~\cite{DBLP:conf/cvpr/LiuCLQLW19} & ResNet18 &  1x  & 33.67 & 54.09 & 35.92 & 19.65 & 36.16 & 43.22  \\
				Faster RCNN &$\dagger$TaskAdap~\cite{DBLP:journals/corr/abs-2006-13108}& ResNet18 &  1x  & 35.77 & 55.22 & 38.74 & 19.32 & 38.72 & 47.27 \\
				\multicolumn{1}{>{\columncolor{Gray}[6pt][418pt]}c|}{Faster RCNN}
				&Ours    & ResNet18  &  1x  & \textbf{37.25} & \textbf{57.09} & \textbf{40.48} & \textbf{20.84} & \textbf{39.94} & \textbf{49.61} \\
				Faster RCNN &AttentionGuided~\cite{zhang2021improve} & ResNet18 &  2x  & 37.00 & 57.20&39.70 &19.90 &39.70 &\textbf{50.30}  \\
				\multicolumn{1}{>{\columncolor{Gray}[6pt][418pt]}c|}{Faster RCNN}
				& Ours & ResNet18 &  2x  & \textbf{38.09} & \textbf{58.33} & \textbf{41.26} & \textbf{21.17} & \textbf{41.09} & 50.16  \\
				\midrule
				\midrule
				Faster RCNN &Student Baseline & ResNet50 & 1x  & 38.03 & 58.91&41.13 &22.21 &41.46 &49.22  \\
				Faster RCNN &Teacher Baseline & ResNet101 & 3x  & 42.03 &62.48 &45.87 &25.22 &45.55 &54.59  \\
				\midrule
				Faster RCNN &$\dagger$FGFI~\cite{DBLP:conf/cvpr/WangYZF19} & ResNet50 &  1x  & 38.85 & 59.62& 42.16 & 22.68 & 42.20 & 50.48  \\
				Faster RCNN &$\dagger$PixelPairWise~\cite{DBLP:conf/cvpr/LiuCLQLW19} & ResNet50 &  1x  & 38.29 & 58.47 &41.83 &21.95 &41.67 &49.33  \\
				Faster RCNN &$\dagger$TaskAdap~\cite{DBLP:journals/corr/abs-2006-13108}& ResNet50 &  1x  & 39.89 & 60.03& 43.19 & 23.73 & 43.23 &52.34  \\
				\multicolumn{1}{>{\columncolor{Gray}[6pt][418pt]}c|}{Faster RCNN}
				&Ours    & ResNet50  &  1x  & \textbf{40.57} & \textbf{61.15} & \textbf{44.38} & \textbf{24.17} &\textbf{44.06} &\textbf{52.80}  \\
				Faster RCNN &AttentionGuided~\cite{zhang2021improve} & ResNet50 &  2x  & 41.50 & \textbf{62.20} & 45.10 & 23.50 & 45.00 & \textbf{55.30} \\
				\multicolumn{1}{>{\columncolor{Gray}[6pt][418pt]}c|}{Faster RCNN}
				&Ours    & ResNet50  &  2x  & \textbf{41.55} & 62.15 & \textbf{45.27}&\textbf{24.44} &\textbf{45.34} &53.95  \\
				\midrule
				\midrule
				Faster RCNN &Student Baseline & MNV2 & 1x  & 29.47 &48.87 & 30.90& 38.86& 30.77 & 16.33  \\
				Faster RCNN &Teacher Baseline & ResNet50 & 3x  & 40.22 & 61.01&43.81 &24.15 &43.52 &51.97  \\
				\midrule
				Faster RCNN &$\dagger$FGFI~\cite{DBLP:conf/cvpr/WangYZF19} & MNV2 &  1x  & 30.27 & 49.87 & 31.60 & 17.03 & 31.82 & 40.06  \\
				Faster RCNN &$\dagger$PixelPairWise~\cite{DBLP:conf/cvpr/LiuCLQLW19} & MNV2 &  1x  & 31.52 & 50.72 & 33.35 & 17.66 & 33.52 & 40.75  \\
				Faster RCNN &$\dagger$TaskAdap~\cite{DBLP:journals/corr/abs-2006-13108}& MNV2 &  1x  & 31.90 & 50.54& 34.26 &16.92 &33.46 &42.82  \\
				\multicolumn{1}{>{\columncolor{Gray}[6pt][418pt]}c|}{Faster RCNN}
				&Ours    & MNV2  &  1x  & \textbf{34.44} & \textbf{53.85} & \textbf{37.04} & \textbf{18.53} & \textbf{36.30} & \textbf{46.92}  \\
				\midrule
				\midrule
				RetinaNet &Student Baseline & ResNet18 & 1x & 31.60 & 49.61 & 33.36 & 17.06 &34.80 &41.11  \\
				RetinaNet &Teacher Baseline & ResNet50 & 3x  & 38.67 &57.99 &41.48 & 23.34&42.30 & 50.31 \\
				\midrule
				RetinaNet &$\dagger$PixelPairWise~\cite{DBLP:conf/cvpr/LiuCLQLW19} & ResNet18 &  1x  &32.48 &50.66 &33.86 &17.30 &35.82 &42.71  \\
				\multicolumn{1}{>{\columncolor{Gray}[6pt][418pt]}c|}{RetinaNet}
				&Ours & ResNet18  &  1x  & \textbf{34.72} & \textbf{53.12} & \textbf{36.73} &\textbf{19.41} &\textbf{38.05} &\textbf{45.93}  \\
				RetinaNet &AttentionGuided~\cite{zhang2021improve} & ResNet18 &  2x  &35.90 &54.40 &38.00 & 17.90&39.10 &\textbf{49.40} \\
				\multicolumn{1}{>{\columncolor{Gray}[6pt][418pt]}c|}{RetinaNet}
				&Ours    & ResNet18  &  2x  & \textbf{36.78}& \textbf{55.35} &\textbf{38.98} &\textbf{20.61} &\textbf{40.35} &47.84  \\
				\bottomrule
			\end{tabular}
		\end{center}
		\caption{Object detection results Box AP, \vs state-of-the-art method on COCO2017 \texttt{val}.} 
		\label{table:comparesota}
	\end{table*}

	We further validate our proposed method on the COCO dataset~\cite{DBLP:conf/eccv/LinMBHPRDZ14} and compare with recent state-of-the-art methods using Faster R-CNN and RetinaNet student-teacher distillation pairs. Results are presented in Table~\ref{table:comparesota} for fast 1x schedule training, since \cite{zhang2021improve} only has 2x schedule results, so we add extra 2x schedule experiments. We don't compare ours with \cite{DBLP:conf/cvpr/WangYZF19} and \cite{DBLP:journals/corr/abs-2006-13108} on RetinaNet because their methods cannot be utilized in one-stage detector. 
	
	Results shows that our method outperforms all previous methods by a large margin with heterogeneous student-teacher backbones and training schedules on both Faster R-CNN and RetinaNet. 
	Surprisingly, our method surpasses the pixel pairwise distillation method~\cite{DBLP:conf/cvpr/LiuCLQLW19} on four distillation pairs by 2.68 AP on average, indicating that the distillation of instance relations makes more difference than pixel relations which is designed for semantic segmentation in detection task. Also, especially in smaller models, our method improves state-of-the-art~\cite{zhang2021improve} by 1.1/0.9 AP on Faster R-CNN and RetinaNet respectively in 2x schedule, even though we didn't add extra parametric modules.

	\subsection{Experiments for Instance Segmentation}
	
	Our distillation framework can be easily extended to the instance segmentation task. We adopt Mask R-CNN~\cite{DBLP:conf/iccv/HeGDG17} as our architecture and evaluate two student-teacher pairs~(R18-R50, R50-R101). Models are trained on COCO2017 images that contain annotated masks, and we report the standard evaluation metric AP$_{box}$ and AP$_{mask}$ based on \textit{Box} IoU and \textit{Mask} IoU respectively. All other training setting is the same as that described in Section \ref{section:training_detail}.  
	
	Results are shown in Table~\ref{table:cocoperformance_instanceseg}. Distilled via our method, Mask R-CNN with ResNet18 surpasses the PixelPairWise~\cite{DBLP:conf/cvpr/LiuCLQLW19} by 3.47 point AP$_{mask}$. In larger backbones, distilled Mask R-CNN with ResNet50 improves the state-of-the-art~\cite{DBLP:conf/cvpr/WangYZF19} and \cite{zhang2021improve} by 2.41/0.12 AP$_{mask}$ in 1x/2x training. Similarly, student-3x models exhibit even higher improvement, bringing 3.5 point AP$_{mask}$ gain on average to the student baseline.
	Basically, the gaps shorten in AP$_{mask}$ are less obvious than that in AP$_{box}$, and it is principally caused by the fact that we do not apply our method to mask head.

	\subsection{Visualization of Graph}
	\begin{figure*}[t]
		\begin{center}
			\subfigure[Labeled Image]{
				\begin{minipage}[b]{0.25\linewidth}
					\centering
					\includegraphics[height=0.621\linewidth]{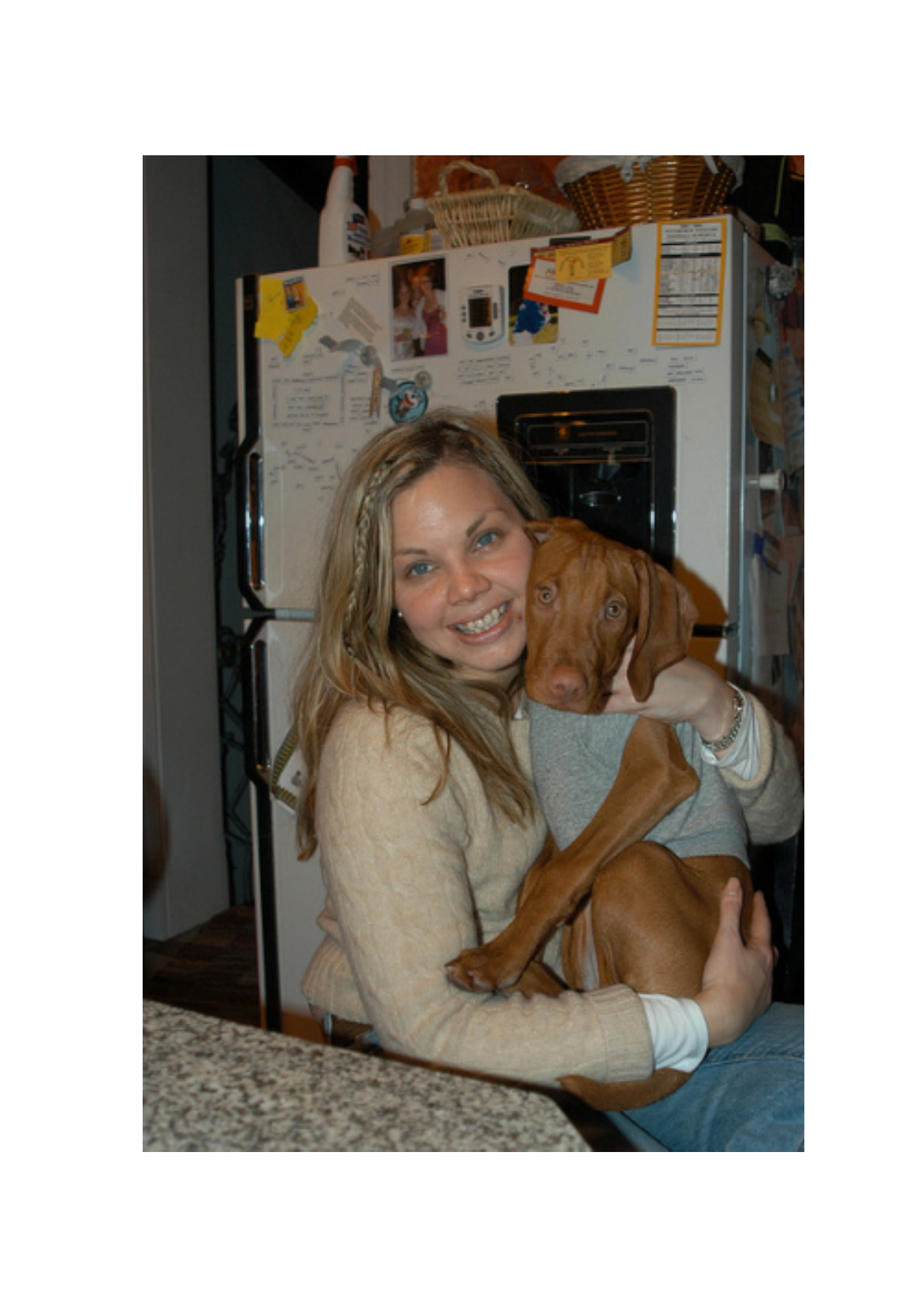}\\\vspace{2pt}
					\includegraphics[height=0.621\linewidth]{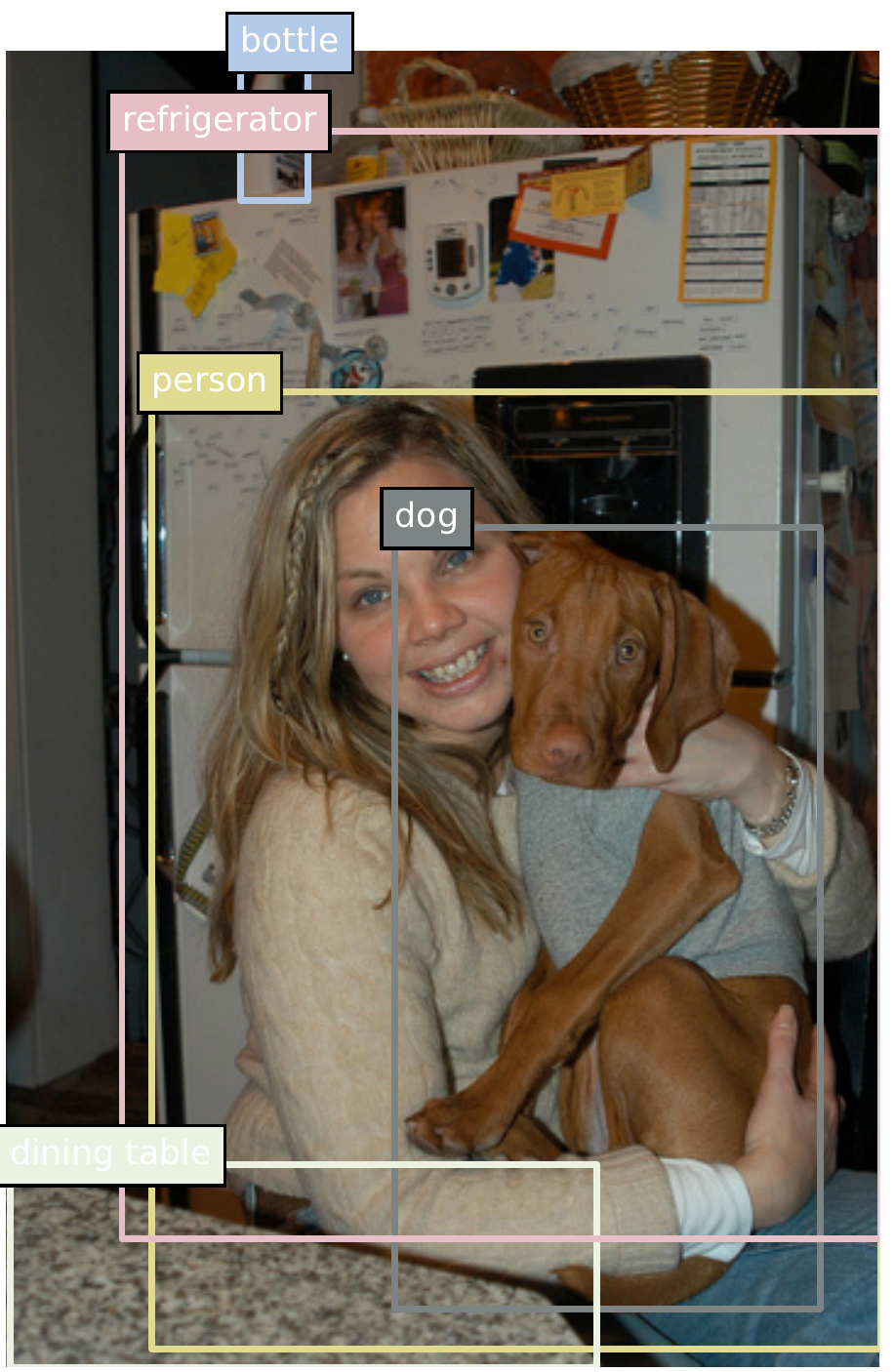}\vspace{2pt}
				\end{minipage}
			}\hspace{-1.1cm}
			\subfigure[Node Visualization]{\label{fig:nodeviz}
				\begin{minipage}[b]{0.25\linewidth}
					\centering
					\includegraphics[width=0.9\linewidth, height=0.63\linewidth]{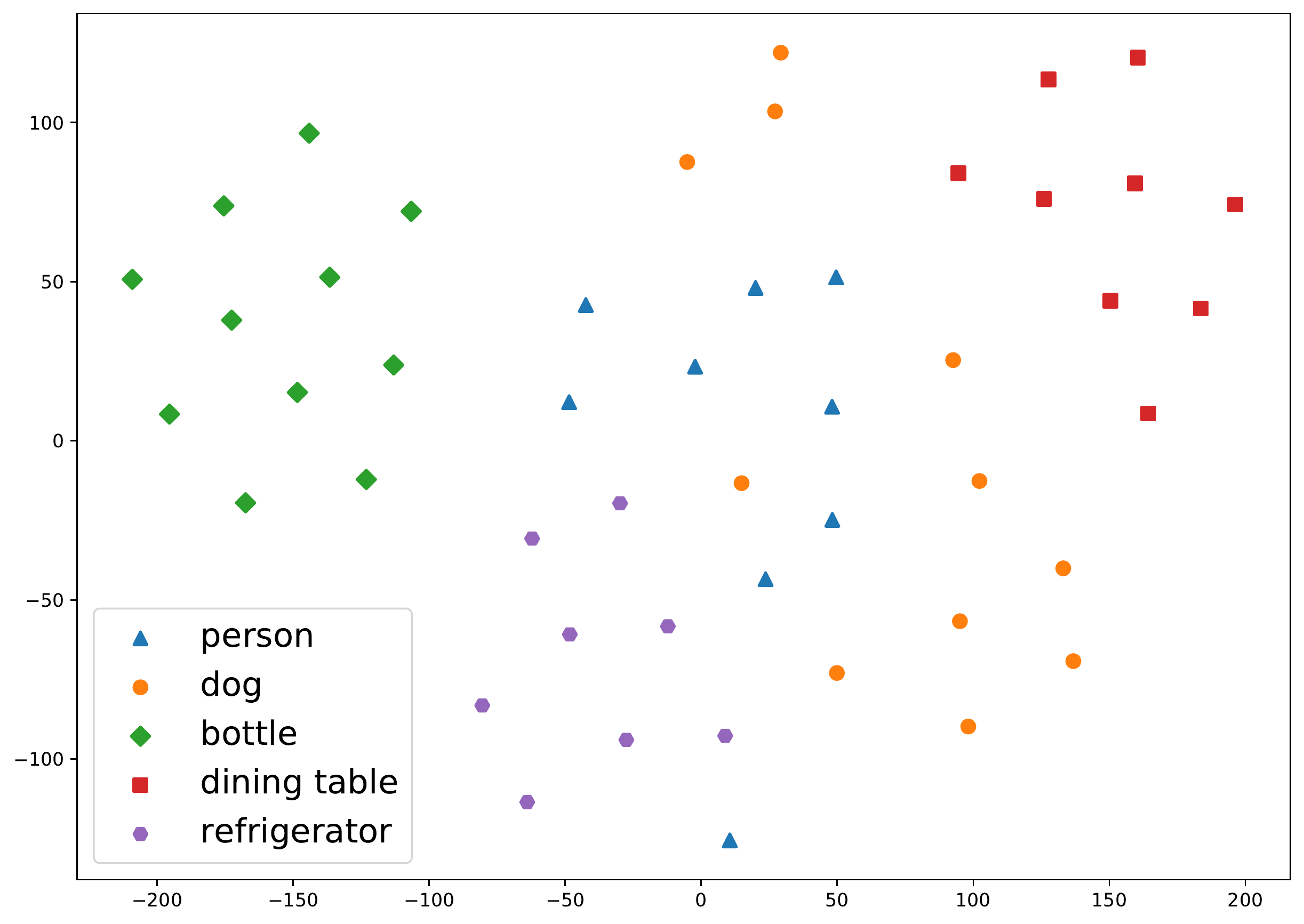}\\
					\includegraphics[width=0.9\linewidth, height=0.63\linewidth]{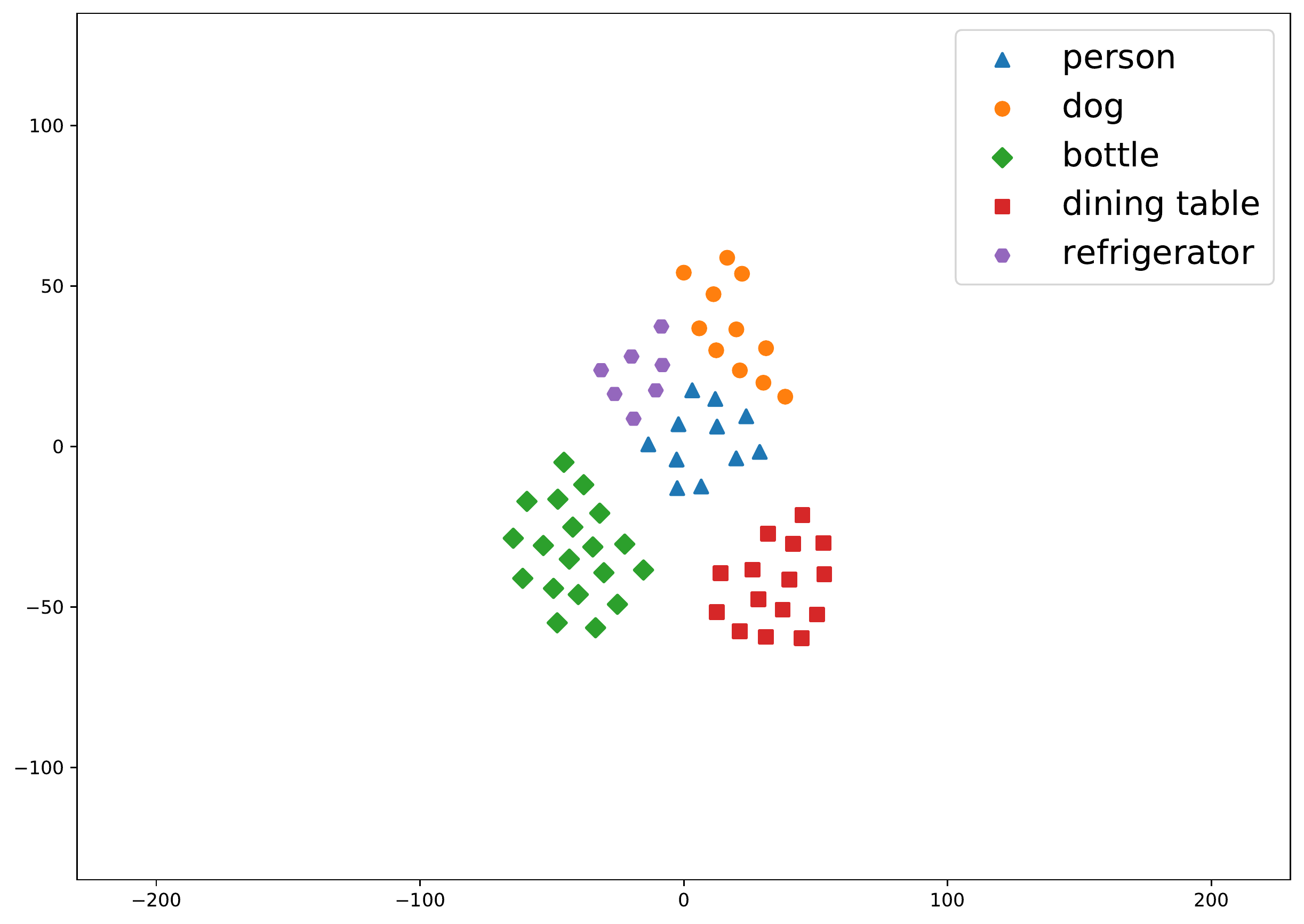}
				\end{minipage}
			}\hspace{-0.25cm} 
			\subfigure[Edge Visualization]{\label{fig:edgeviz}
				\begin{minipage}[b]{0.25\linewidth}
					\centering
					\includegraphics[width=0.9\linewidth, height=0.63\linewidth]{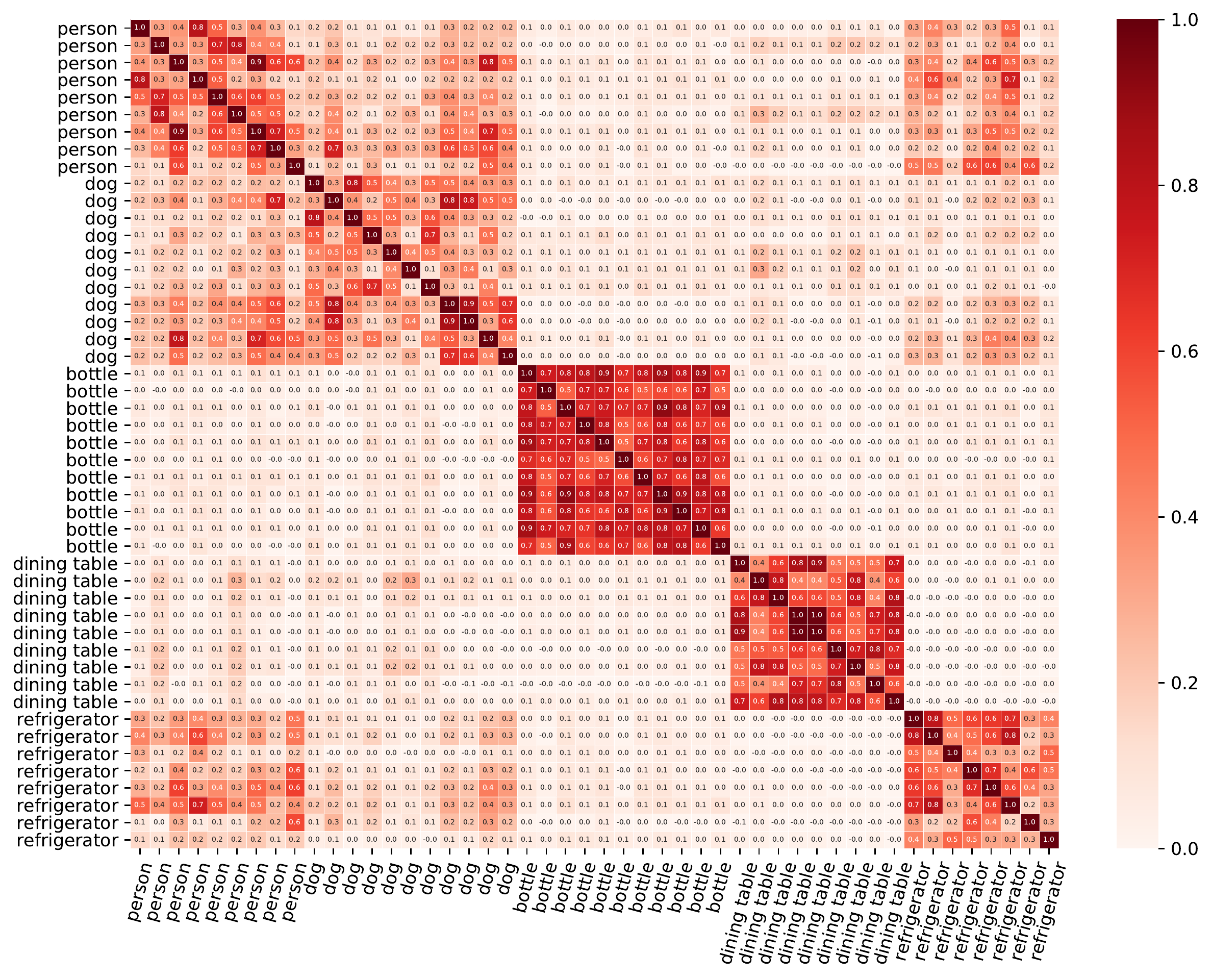}\\
					\includegraphics[width=0.9\linewidth, height=0.63\linewidth]{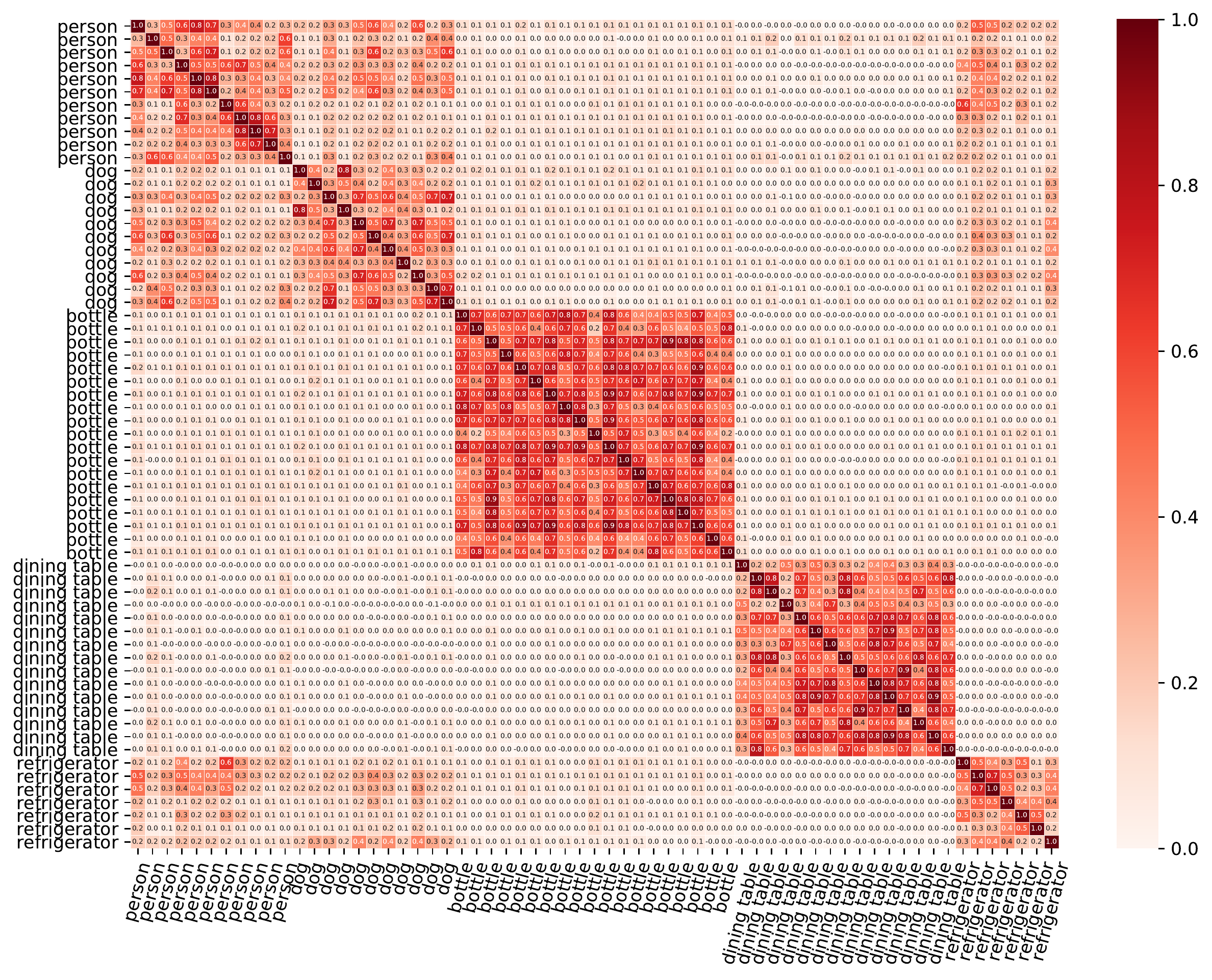}
				\end{minipage}
			}\hspace{-0.25cm}
			\subfigure[AP \vs Edge distance in training]{\label{fig:edgeandap}
				\begin{minipage}[b]{0.25\linewidth}
					\centering
					\includegraphics[width=0.9\linewidth, height=0.63\linewidth]{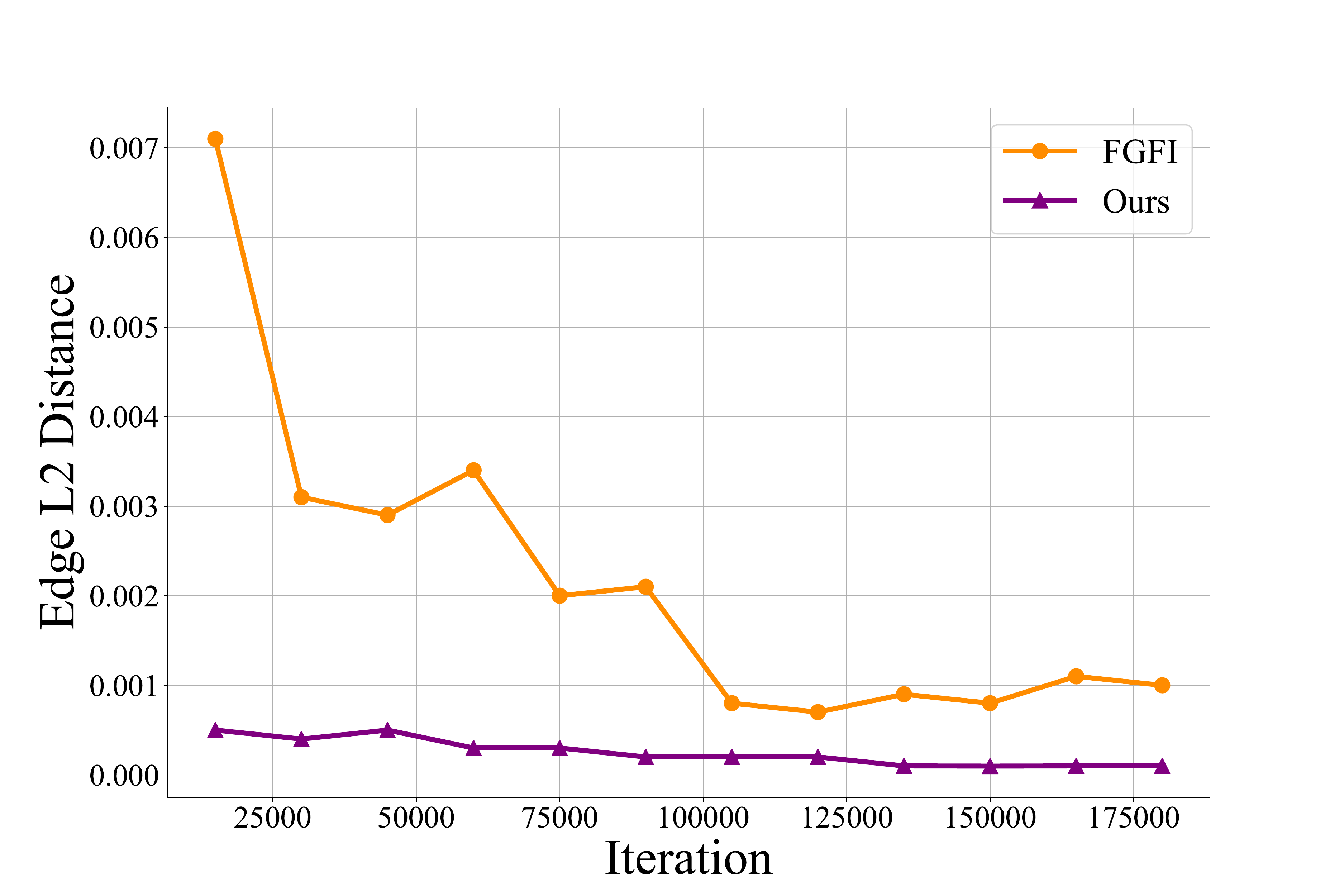}\\
					\includegraphics[width=0.9\linewidth, height=0.63\linewidth]{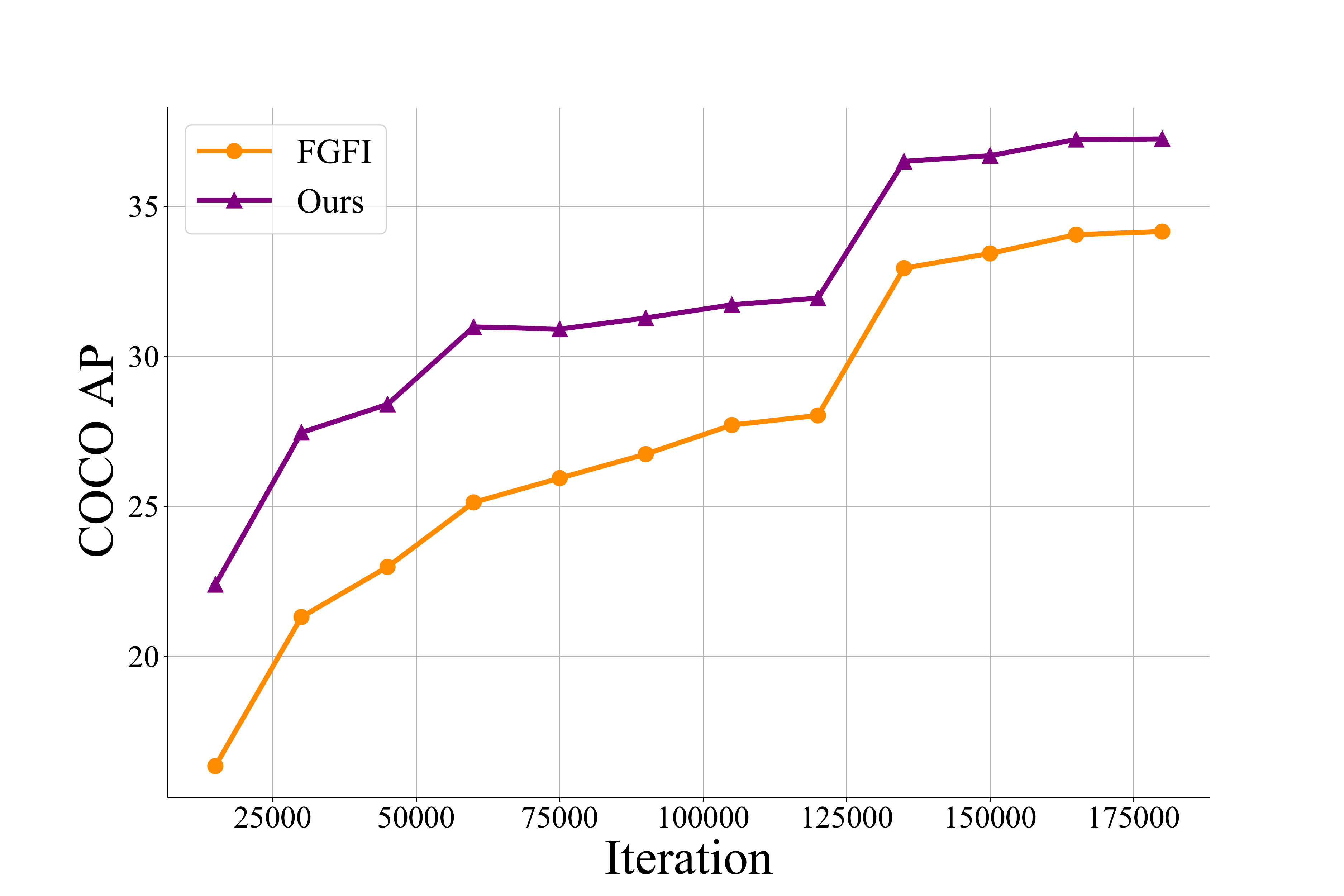}
				\end{minipage}
			}
		\end{center}
		\caption{Graph visualization on test images. The top row of (b)\&(c) represents student results while the bottom row is with teacher results. In (b), we adopt t-SNE~\cite{vanDerMaaten2008} to project high-dimensional node features to 2D space -- each marker represents one node. In (c), we visualize edges as a symmetric matrix by heatmap. The darker matrix element is, the closer relation between two corresponding nodes have~(best view after zoom-in). In (d), we quantitatively compare the edges distance and detection performance of FGFI~\cite{DBLP:conf/cvpr/WangYZF19} and ours. }
		\label{fig:viz}
	\end{figure*}

	
	To better understand how a structured graph manages exploited deep knowledge, we visualize the structured graph from trained student/teacher Faster R-CNN detector. Results are shown in Figure~\ref{fig:viz}. We visualize one image from the COCO dataset. It is observable that the graph nodes extracted from the trained teacher~(Figure~\ref{fig:nodeviz} bottom) are well-clustered in embedding space. However, for the student~(Figure~\ref{fig:nodeviz} top), the nodes labeled as \textit{person} are mixed with those with label \textit{dog} -- these nodes indeed scatter compared with the teacher. 
	
	\begin{table*}[t]
		\begin{center}
			\renewcommand\arraystretch{0.98}
			\begin{tabular}{cccc|cccccc}
				\toprule
				STU & EDG & FGN & BGN & AP & AP$_{50}$ & AP$_{75}$ & AP$_S$ & AP$_M$ & AP$_L$  \\
				\midrule \toprule
				\checkmark &  &  &  & 33.06 & 53.43 & 35.19 & 18.83 & 35.64 & 42.73 \\
				\checkmark & \checkmark &  &  & 33.95 & 53.81 & 36.54 & 18.56 & 36.72 & 44.16 \\
				\checkmark & \checkmark & \checkmark &  & 36.64 & 56.89 & 39.60 & \textbf{21.21} & 39.47 & 48.43 \\
				\checkmark & \checkmark & \checkmark & \checkmark & \textbf{37.17} & \textbf{57.36} & \textbf{40.17} & 21.05 & \textbf{39.97} & \textbf{48.63} \\
				\bottomrule
				\multicolumn{4}{c|}{Teacher} & 40.22 & 61.01 & 43.81 & 24.15 & 43.52 & 51.97 \\
				\bottomrule
			\end{tabular}
		\end{center}
		\caption{Ablations. We adopt R18-R50 student-teacher pair trained on COCO2017 \texttt{train} and tested on COCO2017 \texttt{val}. We conduct Student Baseline~(\textbf{STU}) and gradually add Edge~(\textbf{EDG}), ForeGround Node~(\textbf{FGN}), BackGround Node~(\textbf{BGN}).}
		\vspace{-0.1in}
		\label{table:ablation_r18r50}
	\end{table*}
	For edges, the similarities are much closer within the same classes and are more discriminative in different classes. Obviously, the \textit{person}, \textit{refrigerator}, \textit{dining table}, and \textit{dog} nodes exhibit relatively close inter-class relation, mainly due to the fact that these nodes' features share highly-overlapped regions. However, a good detector should be able to detect largely occluded objects. In teacher, some edges~(Figure~\ref{fig:edgeviz} bottom) exhibit weak intensity~(\textit{person}$\leftrightarrow$\textit{refrigerator} and \textit{dining table}$\leftrightarrow$\textit{person}\&\textit{dog}). But the counterparts~(Figure~\ref{fig:edgeviz} top) in student still have strong links, which make them hard to be correctly classified. These two phenomena exhibited in the visualization further demonstrate the necessity of our method to structurally distill knowledge. We show more COCO examples in our supplementary material.
	
	In Figure~\ref{fig:edgeandap}, to compare ours with pixel-pixel method FGFI~\cite{DBLP:conf/cvpr/WangYZF19} quantitatively, we adopt risk function to evaluate the discrepancy between edges produced by them and teacher as $D(\mathcal{E}^t, \mathcal{E}^s) = \mathbb{E}_{e_{i,j} \sim \mathcal{E}}{\|e_{i,j}^t - e_{i,j}^s\|^2}$ during training, along with the detection performance AP on the COCO benchmark. Obviously, without distilling pairwise relation, the edge distance gap still remains too large between FGFI and the teacher, while our method achieves nearly \textbf{0} distance towards teacher, resulting in substantial improvement in terms of COCO AP than the pixel-to-pixel scheme.   
	
	\subsection{Ablation Study}
	
	As shown in Table~\ref{table:ablation_r18r50}, we conduct experiments on different combinations of components for graph distillation to highlight that each part of our proposed method makes difference. We have three different modules contributing to graph distillation loss in our framework. They are 1) edge, 2) foreground node, and 3) background node.

	\vspace{-0.1in}
	\paragraph{Edges}
	Preserving the same edge structures between students and teachers contributes 0.89 point AP to distillation performance. It indicates that even without straightforward pixel-pixel mimicking, merely aligning relations can be an essential regularization to preserve topological structure, which proves that our method is feasible.      
	
	\vspace{-0.1in}
	\paragraph{Foreground Nodes}
	Imitating student features in foreground-labeled nodes brings about 2.69 AP gain, which is greater than that from edges, it means that distilling foreground features effectively enables the student networks to focus more on the regions of foreground instances. This suggests that features matching in these foreground-labeled areas is more salient for the student to imitate than the global high-dimensional feature maps without much noise. Moreover, edges cooperating with nodes yield more promising results, which verifies the effectiveness of both parts of the graph -- they are complementary.       
	
	\vspace{-0.1in}
	\paragraph{Background Nodes}
	Adding imitation of student features in background-labeled regions brings extra AP gain, which is 0.53 compared to foreground nodes. This suggests that, even on the basis of foreground nodes imitation, seemingly useless background nodes play an important role in distilling students when balanced via our adaptive background loss weight.
	
	\section{Conclusion}
	In this paper, we have proposed a new \textit{Structured Instance Graph} to manage instances in the detection distillation system. 
	We adopt it to leverage useful local proposal-level features while maintaining their global semantic inter-relations for distillation.
	Extensive experiments are conducted to manifest the effectiveness and robustness of distilling the whole structured graph regarding both object detection and instance segmentation distillation tasks. 
	
	{\small
		\bibliographystyle{ieee_fullname}
		\bibliography{ICCV_2021}

\begin{thebibliography}{10}\itemsep=-1pt

\bibitem{Cai_2019}
Zhaowei Cai and Nuno Vasconcelos.
\newblock Cascade r-cnn: High quality object detection and instance
  segmentation.
\newblock {\em TPAMI}, 2019.

\bibitem{DBLP:conf/nips/ChenCYHC17}
Guobin Chen, Wongun Choi, Xiang Yu, Tony~X. Han, and Manmohan Chandraker.
\newblock Learning efficient object detection models with knowledge
  distillation.
\newblock In {\em NIPS}, 2017.

\bibitem{DBLP:conf/nips/DaiLHS16}
Jifeng Dai, Yi Li, Kaiming He, and Jian Sun.
\newblock {R-FCN:} object detection via region-based fully convolutional
  networks.
\newblock In {\em NIPS}, 2016.

\bibitem{girshickICCV15fastrcnn}
Ross Girshick.
\newblock Fast {R-CNN}.
\newblock In {\em ICCV}, 2015.

\bibitem{DBLP:conf/cvpr/GirshickDDM14}
Ross Girshick, Jeff Donahue, Trevor Darrell, and Jitendra Malik.
\newblock Rich feature hierarchies for accurate object detection and semantic
  segmentation.
\newblock In {\em CVPR}, 2014.

\bibitem{DBLP:journals/corr/HanMD15}
Song Han, Huizi Mao, and William~J. Dally.
\newblock Deep compression: Compressing deep neural network with pruning,
  trained quantization and huffman coding.
\newblock In {\em ICLR}, 2016.

\bibitem{DBLP:conf/nips/HanPTD15}
Song Han, Jeff Pool, John Tran, and William~J. Dally.
\newblock Learning both weights and connections for efficient neural network.
\newblock In {\em NIPS}, 2015.

\bibitem{DBLP:conf/iccv/HeGDG17}
Kaiming He, Georgia Gkioxari, Piotr Doll{\'{a}}r, and Ross Girshick.
\newblock Mask {R-CNN}.
\newblock In {\em ICCV}, 2017.

\bibitem{DBLP:conf/eccv/HeZR014}
Kaiming He, Xiangyu Zhang, Shaoqing Ren, and Jian Sun.
\newblock Spatial pyramid pooling in deep convolutional networks for visual
  recognition.
\newblock In {\em ECCV}, 2014.

\bibitem{DBLP:conf/cvpr/HeZRS16}
Kaiming He, Xiangyu Zhang, Shaoqing Ren, and Jian Sun.
\newblock Deep residual learning for image recognition.
\newblock In {\em CVPR}, 2016.

\bibitem{Hinton2015DistillingTK}
Geoffrey~E. Hinton, Oriol Vinyals, and Jeffrey Dean.
\newblock Distilling the knowledge in a neural network.
\newblock {\em ArXiv}, abs/1503.02531, 2015.

\bibitem{DBLP:conf/cvpr/HuGZDW18}
Han Hu, Jiayuan Gu, Zheng Zhang, Jifeng Dai, and Yichen Wei.
\newblock Relation networks for object detection.
\newblock In {\em CVPR}, 2018.

\bibitem{DBLP:conf/cvpr/JacobKCZTHAK18}
Benoit Jacob, Skirmantas Kligys, Bo Chen, Menglong Zhu, Matthew Tang, Andrew~G.
  Howard, Hartwig Adam, and Dmitry Kalenichenko.
\newblock Quantization and training of neural networks for efficient
  integer-arithmetic-only inference.
\newblock In {\em CVPR}, 2018.

\bibitem{DBLP:conf/nips/LanZG18}
Xu Lan, Xiatian Zhu, and Shaogang Gong.
\newblock Knowledge distillation by on-the-fly native ensemble.
\newblock In {\em NIPS}, 2018.

\bibitem{DBLP:conf/cvpr/LiJY17}
Quanquan Li, Shengying Jin, and Junjie Yan.
\newblock Mimicking very efficient network for object detection.
\newblock In {\em CVPR}, 2017.

\bibitem{DBLP:conf/cvpr/LiWLQYF19}
Rundong Li, Yan Wang, Feng Liang, Hongwei Qin, Junjie Yan, and Rui Fan.
\newblock Fully quantized network for object detection.
\newblock In {\em CVPR}, 2019.

\bibitem{DBLP:conf/cvpr/LinDGHHB17}
Tsung{-}Yi Lin, Piotr Doll{\'{a}}r, Ross Girshick, Kaiming He, Bharath
  Hariharan, and Serge~J. Belongie.
\newblock Feature pyramid networks for object detection.
\newblock In {\em CVPR}, 2017.

\bibitem{DBLP:conf/eccv/LinMBHPRDZ14}
Tsung{-}Yi Lin, Michael Maire, Serge~J. Belongie, James Hays, Pietro Perona,
  Deva Ramanan, Piotr Doll{\'{a}}r, and C.~Lawrence Zitnick.
\newblock Microsoft {COCO:} common objects in context.
\newblock In {\em ECCV}, 2014.

\bibitem{lin2017focal}
Tsung-Yi Lin, Priya Goyal, Ross Girshick, Kaiming He, and Piotr Doll{\'a}r.
\newblock Focal loss for dense object detection.
\newblock In {\em ICCV}, 2017.

\bibitem{Liu_2016}
Wei Liu, Dragomir Anguelov, Dumitru Erhan, Christian Szegedy, Scott Reed,
  Cheng-Yang Fu, and Alexander~C. Berg.
\newblock {SSD:} single shot multibox detector.
\newblock {\em ECCV}, 2016.

\bibitem{DBLP:conf/cvpr/LiuCLQLW19}
Yifan Liu, Ke Chen, Chris Liu, Zengchang Qin, Zhenbo Luo, and Jingdong Wang.
\newblock Structured knowledge distillation for semantic segmentation.
\newblock In {\em CVPR}, 2019.

\bibitem{DBLP:conf/cvpr/Liu0SC18}
Yong Liu, Ruiping Wang, Shiguang Shan, and Xilin Chen.
\newblock Structure inference net: Object detection using scene-level context
  and instance-level relationships.
\newblock In {\em CVPR}, 2018.

\bibitem{DBLP:conf/cvpr/ParkKLC19}
Wonpyo Park, Dongju Kim, Yan Lu, and Minsu Cho.
\newblock Relational knowledge distillation.
\newblock In {\em CVPR}, 2019.

\bibitem{DBLP:conf/eccv/RastegariORF16}
Mohammad Rastegari, Vicente Ordonez, Joseph Redmon, and Ali Farhadi.
\newblock Xnor-net: Imagenet classification using binary convolutional neural
  networks.
\newblock In {\em ECCV}, 2016.

\bibitem{DBLP:conf/cvpr/RedmonDGF16}
Joseph Redmon, Santosh~Kumar Divvala, Ross Girshick, and Ali Farhadi.
\newblock You only look once: Unified, real-time object detection.
\newblock In {\em CVPR}, 2016.

\bibitem{renNIPS15fasterrcnn}
Shaoqing Ren, Kaiming He, Ross Girshick, and Jian Sun.
\newblock Faster {R-CNN}: Towards real-time object detection with region
  proposal networks.
\newblock In {\em NIPS}, 2015.

\bibitem{DBLP:journals/corr/RomeroBKCGB14}
Adriana Romero, Nicolas Ballas, Samira~Ebrahimi Kahou, Antoine Chassang, Carlo
  Gatta, and Yoshua Bengio.
\newblock Fitnets: Hints for thin deep nets.
\newblock In {\em ICLR}, 2015.

\bibitem{DBLP:conf/cvpr/SandlerHZZC18}
Mark Sandler, Andrew~G. Howard, Menglong Zhu, Andrey Zhmoginov, and
  Liang{-}Chieh Chen.
\newblock Mobilenetv2: Inverted residuals and linear bottlenecks.
\newblock In {\em CVPR}, 2018.

\bibitem{DBLP:conf/cvpr/ShrivastavaGG16}
Abhinav Shrivastava, Abhinav Gupta, and Ross~B. Girshick.
\newblock Training region-based object detectors with online hard example
  mining.
\newblock In {\em CVPR}, 2016.

\bibitem{DBLP:journals/corr/SimonyanZ14a}
Karen Simonyan and Andrew Zisserman.
\newblock Very deep convolutional networks for large-scale image recognition.
\newblock In {\em ICLR}, 2015.

\bibitem{DBLP:journals/corr/abs-2006-13108}
Ruoyu Sun, Fuhui Tang, Xiaopeng Zhang, Hongkai Xiong, and Qi Tian.
\newblock Distilling object detectors with task adaptive regularization.
\newblock {\em CoRR}, 2020.

\bibitem{DBLP:conf/icml/TanL19}
Mingxing Tan and Quoc~V. Le.
\newblock Efficientnet: Rethinking model scaling for convolutional neural
  networks.
\newblock In Kamalika Chaudhuri and Ruslan Salakhutdinov, editors, {\em ICML},
  2019.

\bibitem{DBLP:conf/iccv/TianSCH19}
Zhi Tian, Chunhua Shen, Hao Chen, and Tong He.
\newblock {FCOS:} fully convolutional one-stage object detection.
\newblock In {\em ICCV}, 2019.

\bibitem{vanDerMaaten2008}
Laurens van~der Maaten and Geoffrey Hinton.
\newblock Visualizing data using {t-SNE}.
\newblock {\em Journal of Machine Learning Research}, 2008.

\bibitem{DBLP:conf/cvpr/WangYZF19}
Tao Wang, Li Yuan, Xiaopeng Zhang, and Jiashi Feng.
\newblock Distilling object detectors with fine-grained feature imitation.
\newblock In {\em CVPR}, 2019.

\bibitem{DBLP:conf/cvpr/0004GGH18}
Xiaolong Wang, Ross~B. Girshick, Abhinav Gupta, and Kaiming He.
\newblock Non-local neural networks.
\newblock In {\em CVPR}, 2018.

\bibitem{DBLP:conf/cvpr/0008CYLWL020}
Yue Wu, Yinpeng Chen, Lu Yuan, Zicheng Liu, Lijuan Wang, Hongzhi Li, and Yun
  Fu.
\newblock Rethinking classification and localization for object detection.
\newblock In {\em CVPR}, 2020.

\bibitem{wu2019detectron2}
Yuxin Wu, Alexander Kirillov, Francisco Massa, Wan-Yen Lo, and Ross Girshick.
\newblock Detectron2.
\newblock \url{https://github.com/facebookresearch/detectron2}, 2019.

\bibitem{Zagoruyko2016PayingMA}
Sergey Zagoruyko and Nikos Komodakis.
\newblock Paying more attention to attention: Improving the performance of
  convolutional neural networks via attention transfer.
\newblock In {\em ICLR}, 2016.

\bibitem{zhang2021improve}
Linfeng Zhang and Kaisheng Ma.
\newblock Improve object detection with feature-based knowledge distillation:
  Towards accurate and efficient detectors.
\newblock In {\em ICLR}, 2021.

\bibitem{DBLP:conf/cvpr/ZhangXHL18}
Ying Zhang, Tao Xiang, Timothy~M. Hospedales, and Huchuan Lu.
\newblock Deep mutual learning.
\newblock In {\em CVPR}, 2018.

\bibitem{DBLP:journals/corr/ZhouNZWWZ16}
Shuchang Zhou, Zekun Ni, Xinyu Zhou, He Wen, Yuxin Wu, and Yuheng Zou.
\newblock Dorefa-net: Training low bitwidth convolutional neural networks with
  low bitwidth gradients.
\newblock {\em CoRR}, 2016.

\end{thebibliography}
	}
	
\end{document}